\documentclass[letterpaper]{article}
\usepackage[preprint]{aaai2027}
\usepackage[hyphens]{url}
\usepackage{graphicx}
\usepackage{natbib}
\usepackage{caption}
\usepackage{algorithm}
\usepackage{algorithmic}

\usepackage{newfloat}
\usepackage{listings}
\DeclareCaptionStyle{ruled}{labelfont=normalfont,labelsep=colon,strut=off}
\floatstyle{ruled}
\newfloat{listing}{tb}{lst}{}
\floatname{listing}{Listing}

\usepackage{booktabs}

\newcommand{\keywords}[1]{}

\usepackage{amsmath,amsfonts,bm}

\def\eqref#1{equation~\ref{#1}}

\def\1{\bm{1}}

\DeclareMathAlphabet{\mathsfit}{\encodingdefault}{\sfdefault}{m}{sl}
\SetMathAlphabet{\mathsfit}{bold}{\encodingdefault}{\sfdefault}{bx}{n}

\usepackage{tabularx}
\usepackage{multirow}
\usepackage{xcolor}
\usepackage{colortbl}
\usepackage{subcaption}
\usepackage{makecell}
\usepackage{kotex}
\usepackage{cleveref}
\usepackage{xspace}

\makeatletter
\DeclareRobustCommand\onedot{\futurelet\@let@token\@onedot}
\def\@onedot{\ifx\@let@token.\else.\null\fi\xspace}
\def\eg{\emph{e.g}\onedot}

\makeatother

\definecolor{first}{HTML}{FFCCC9}
\definecolor{second}{HTML}{FFFFC7}

\definecolor{linkcolor}{HTML}{0B5394}
\newcommand{\weburl}[1]{%
    \pdfstartlink attr{/Border[0 0 0]}%
        user{/Subtype/Link/A<</Type/Action/S/URI/URI(#1)>>}%
    \textcolor{linkcolor}{#1}\pdfendlink}

\makeatletter
\g@addto@macro\@maketitle{%
  \begin{center}
    \includegraphics[width=\textwidth, trim={0 0 0 0},clip]{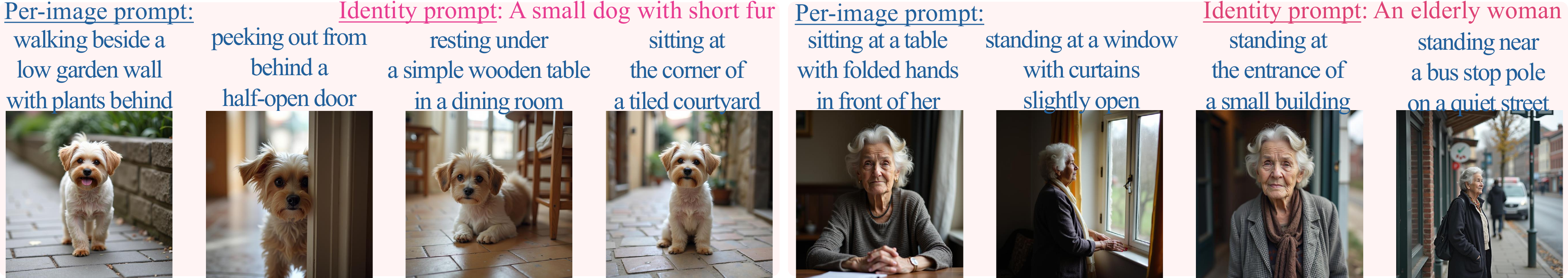}
    \captionof{figure}{Overview of the consistent story generation task and our results. The identity prompt provides the shared character description across all images, while the per-image descriptions specify unique attributes for each image. Our results show consistent character representation across images, with each per-image description effectively reflected.}
    \label{fig:teaser}
  \end{center}
  \vskip 0.5em
}
\makeatother

\title{AsemConsist: Adaptive Semantic Feature Control \\ for Training-Free Identity-Consistent Generation}

\author{
    Shin Seong Kim\equalcontrib,
    Minjung Shin\equalcontrib,
    Hyunin Cho,
    Youngjung Uh\corresponding
}
\affiliations{
    Yonsei University\\
    \{tltydl2, smj139052, hyunin9528, yj.uh\}@yonsei.ac.kr
}

\begin{document}

\maketitle

\begin{abstract}
Recent text-to-image diffusion models have significantly improved visual quality and text alignment. However, generating a sequence of images while preserving consistent character identity across diverse scenes remains challenging. Existing methods often face a trade-off between maintaining identity consistency and per-image prompt alignment.
In this paper, we introduce \textit{AsemConsist}, a framework that resolves this trade-off through selective text embedding modification, enabling consistent identity preservation without degrading per-image prompt alignment.
We further analyze the semantic structure of padding embeddings and find that, in multi-encoder backbones, only padding embeddings that retain prompt-related semantics can effectively serve as semantic containers.
Based on this observation, we selectively inject per-image semantics into such padding embeddings while suppressing prompt-irrelevant components.
Additionally, we propose an adaptive feature-sharing strategy that automatically evaluates identity specificity and selectively applies constraints only to ambiguous identity prompts.
Finally, we propose a unified evaluation metric called \textit{SeeSaw}, which measures the balance between identity consistency and per-image alignment while evaluating whether identity and per-image prompts are equally reflected in generated images.
Our method demonstrates superior performance over existing competitors when built upon SD3.5 and FLUX backbones, highlighting its effectiveness across different architectures and text encoders.
\end{abstract}

\begin{links}
    \renewcommand{\link}[2]{\par\textbf{#1} --- \weburl{#2}}
    \link{Project Page}{https://minjung-s.github.io/asemconsist}
\end{links}

\section{Introduction}

Recent text-to-image (T2I) generative models enable high-quality visuals closely aligned with textual prompts~\cite{sd, sdxl, flux, sd3}. Building on this success, \textit{identity-consistent generation} aims to create a coherent image sequence that preserves the same character identity across scenes. As illustrated in \cref{fig:teaser}, the task fixes a single identity prompt that specifies the shared character and pairs it with a separate per-image prompt for each scene, so that the identity is preserved while every image realizes its own attributes. This capability is increasingly important for narrative-driven content, animation, and interactive storytelling~\cite{story2board, 1p1s, consistory, characonsist}.
Its demand is evident in commercial image generators such as Nano-Banana and GPT-image\footnote{Being closed-source and paid, commercial models are excluded from our main comparison; qualitative results are in \cref{app:commercial}.}, which increasingly support this task.

A key challenge in this task is balancing identity preservation with per-image prompt alignment.
Attention-based feature sharing~\cite{consistory,storydiffusion} often yields fixed poses or unrealistic shapes, and its DiT-based~\cite{dit} successors still lose identity under generic prompts (e.g., ``a man'')~\cite{characonsist} or suffer structural distortions~\cite{codi}.
A complementary line of work adopts the \emph{single-prompt setting}~\cite{1p1s,zigzag,decorstory}, which encodes the identity prompt and all per-image prompts as one sequence for more cohesive identity semantics.
However, existing methods apply a single uniform adjustment to each per-image prompt's embedding, treating it as one unit rather than controlling its individual components~\cite{1p1s,zigzag}.
None asks \emph{which components} of an embedding actually carry the semantics worth amplifying or suppressing.

This question is the starting point of our work.
Our analysis (\cref{sec:analysis}) shows that each per-image embedding \emph{entangles identity-supporting and identity-conflicting semantics}: some of its spectral components reinforce the shared identity, while others conflict with it.
Modifying a per-image embedding as one unit cannot control its identity-supporting and identity-conflicting components separately, so any change to a per-image prompt inevitably affects both. This is exactly the trade-off observed in prior work.
We therefore propose \textit{AsemConsist}, a training-free framework that resolves this trade-off by answering three questions: \emph{which} embedding components to control, \emph{where} to write additional semantics, and \emph{when} image-level constraints are needed.

\emph{Which components?}
We decompose each embedding into spectral components and derive reference vectors from the identity and per-image prompts.
Guided by these, we amplify only the components aligned with both the shared identity and the current per-image description, while suppressing identity-irrelevant ones.
This strengthens identity and per-image semantics simultaneously, without amplifying their entangled residue.

\emph{Where to write semantics?}
We repurpose padding embeddings (\texttt{<pad>}, \texttt{<|endoftext|>}) as \emph{semantic containers}.
Prior work notes that padding embeddings influence generation~\cite{paddingtone,get} but does not characterize them across the heterogeneous text encoders of recent DiT backbones~\cite{sd3,flux}.
By analyzing these encoders, we identify which padding embeddings retain prompt-related semantics and can therefore serve as containers, and write per-image semantics into them.

\emph{When to constrain?}
To keep a character consistent, existing methods reuse image features across generated images to enforce a common appearance~\cite{consistory,characonsist,storydiffusion}. They treat this feature sharing as an always-on mechanism, applying it uniformly to every input, without asking whether a given identity prompt \emph{needs} it. 
We show that it depends on the prompt: specific prompts (e.g., ``a 16 years old girl with wavy blonde hair, a slender frame, and blue eyes'') already pin down appearance, whereas ambiguous ones (e.g., ``a man'') genuinely require an anchor. 
For already-specific prompts, sharing therefore brings no identity gain while needlessly constraining pose and layout diversity.
We propose adaptive feature sharing, which automatically estimates the ambiguity of an identity prompt and engages sharing only when needed, serving as an identity anchor.

Finally, we propose \textit{SeeSaw}, a unified evaluation protocol for identity-consistent generation.
Current metrics score identity preservation and per-image alignment separately, neglecting their balance. Moreover, long identity prompts can dominate alignment scores, masking poor alignment with the per-image descriptions.
SeeSaw jointly measures both aspects and their balance, while checking that identity and per-image prompts are equally represented.

Our main contributions can be summarized as follows:
\begin{itemize}
    \item We trace the identity-alignment trade-off to component-level semantic entanglement, and propose a selective embedding modification that controls individual spectral components rather than each per-image embedding as a whole.
    \item We characterize padding embeddings across FLUX's and SD3.5's text encoders, and exploit those usable as writable semantic containers via selective projection.
    \item We introduce ambiguity-adaptive feature sharing, which constrains image features only when the identity prompt cannot anchor appearance on its own.
    \item We propose \textit{SeeSaw}, a unified metric quantifying identity consistency, per-image alignment, and their balance, while addressing the pitfall of overly long identity prompts.
\end{itemize}

\section{Related Works}

\begin{figure*}[t]
    \centering
    \includegraphics[width=\linewidth]{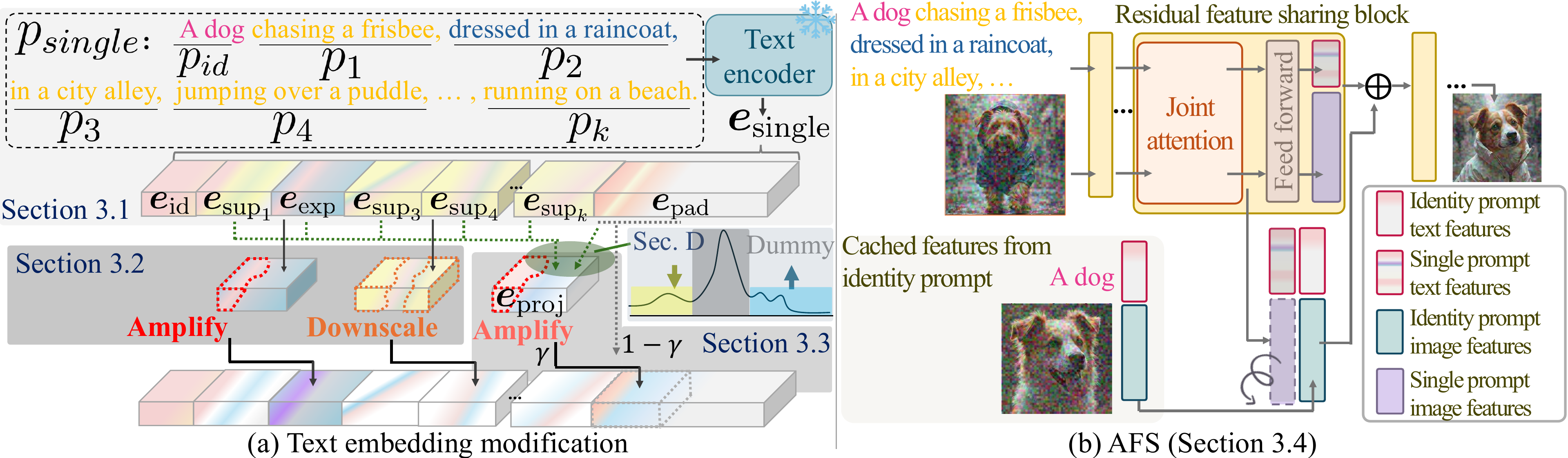}
    \caption{
        Overview of our method.
    }
    \label{fig:overall_method}
\end{figure*}

\subsection{Identity-Consistent Image Generation}
Identity-consistent generation aims to preserve a subject's identity across images while remaining faithful to each prompt~\cite{story2board,fairygen}.
Training-based approaches~\cite{thechosenone,photomaker,pulid,ipadapter,storygptv,storygen,ominicontrol,omnigen,fluxkontext} face restricted domains or an identity-alignment trade-off. Those conditioned on reference images~\cite{ipadapter,ominicontrol,omnigen,fluxkontext} can overfit the given examples, limiting generalization to new appearances or poses.
A common training-free strategy shares image features via attention~\cite{consistory,storydiffusion}. Recent variants adapt it to DiT backbones~\cite{dit} through early-step optimal transport~\cite{codi} or point-tracking attention with fine-grained masks~\cite{characonsist}.
However, these approaches struggle to balance identity consistency with prompt-specific fidelity: uniform sharing constrains pose and layout diversity, masking-based variants~\cite{codi,characonsist} are sensitive to mask quality, and some~\cite{characonsist} perform well mainly when the base model is already consistent.
Another training-free line controls generation in the text embedding space: 1Prompt1Story~\cite{1p1s} encodes the identity prompt and all per-image prompts as one sequence and uniformly rescales the singular values of each per-image embedding.
Later methods build on this joint encoding with multi-phase sampling~\cite{zigzag}, yet all manipulate whole embeddings, rescaling identity-supporting and identity-conflicting semantics together (\cref{sec:analysis}).
In contrast, we control individual spectral components rather than whole embeddings, resolving this trade-off.

\subsection{Metrics for Consistent Image Generation}
Identity preservation and per-image alignment are typically measured separately, using VQAScore~\cite{vqascore} for alignment and DINO or DreamSim~\cite{dino, dreamsim} for consistency.
However, VQA-based scores are right-skewed and saturate near their upper bound~\cite{eval_evaluators}, so long identity prompts can inflate alignment scores even when per-image attributes are missed. Moreover, separate reporting fails to yield a unified ranking.
Our \textit{SeeSaw} metric addresses both issues (\cref{sec:seesaw}).

\section{Method}

In this section, we first describe our problem setting (\cref{sec:problem_setting}).
Next, we introduce our selective text embedding modification, which simultaneously ensures identity consistency and per-image prompt alignment (\emph{which components}; \cref{sec:svd}). We then describe the use of padding embeddings as semantic containers (\emph{where to write}; \cref{sec:semantic_container}).
Finally, we propose an adaptive feature-sharing mechanism that assesses the ambiguity of the identity prompt and applies feature sharing only when needed (\emph{when to constrain}; \cref{sec:ambiguity}).
All strategies operate at inference time without training and apply to both FLUX~\cite{flux} and SD3.5~\cite{sd3}, highlighting their architecture-independence.

\subsection{Problem Setting} \label{sec:problem_setting}

Our goal is to generate a coherent image sequence that preserves the same character identity. Formally, an identity-consistent generation framework takes an identity prompt $p_{\mathrm{id}}$ describing the character to preserve, and per-image prompts $\{p_i\}_{i=1}^{k}$ specifying scene-specific attributes such as pose and background.

We adopt the \emph{single-prompt setting}~\cite{1p1s,zigzag,decorstory}, which encodes the identity prompt and all per-image prompts as one sequence; joint encoding yields more cohesive identity semantics than independent encoding, as shown in \cref{app:single_prompt}.
Concretely, we concatenate the identity prompt and all per-image prompts into a single prompt $p_{\mathrm{single}} = [p_{\mathrm{id}},\, p_1,\, \ldots,\, p_k]$.
Given the combined prompt $p_{\mathrm{single}}$ of length $L_{\mathrm{single}}$, the text encoder $E(\cdot)$ produces a single embedding:
\[
\boldsymbol{e}_{\mathrm{single}} = [\boldsymbol{e}_{\mathrm{id}},\, \boldsymbol{e}_1,\, \ldots,\, \boldsymbol{e}_k] \in \mathbb{R}^{L_{\mathrm{single}} \times d},
\]
where $d$ denotes the embedding dimension and each $\boldsymbol{e}_\ast \in \mathbb{R}^{L_{\mathrm{\ast}} \times d}$ corresponds to its prompt $p_\ast$.
Importantly, since the text encoder mixes information across tokens through attention, each prompt $p_\ast$ influences all embeddings $\boldsymbol{e}_\ast$\footnote{Although our notation $\boldsymbol{e}_\ast$ denotes a matrix consisting of token-wise embedding vectors, we treat it as an embedding for simplicity.}; this global mixing is precisely what our embedding modification in \cref{sec:svd} exploits.

When generating an image corresponding to a specific prompt $p_i$, we denote this prompt as the \emph{expression prompt} $p_{\mathrm{exp}}$ with its embedding $\boldsymbol{e}_{\mathrm{exp}}$, following the standard terminology of the single-prompt setting~\cite{1p1s,zigzag}. The remaining prompts $\{p_j\}_{j \in \{1, \ldots, k\} \setminus \{i\}}$, which should not influence the current image, are termed \emph{suppression prompts}, with corresponding embeddings denoted as ${\boldsymbol{e}_{\mathrm{sup}_j}}$.

As illustrated in \cref{fig:overall_method}(a), to generate an image from the second per-image prompt (e.g., ``dressed in a raincoat''), we form embeddings:
\[
[\boldsymbol{e}_{\mathrm{id}},\, \boldsymbol{e}_{\mathrm{sup}_1},\, \boldsymbol{e}_{\mathrm{exp}},\, \boldsymbol{e}_{\mathrm{sup}_3},\, \dots,\, \boldsymbol{e}_{\mathrm{sup}_k}],
\]
which explicitly express the desired attributes (e.g., the dog wearing a raincoat) while suppressing irrelevant information from other prompts.

\subsection{Selective Text Embedding Modification (STM)} \label{sec:svd}
Under the single-prompt setting, individual embeddings $\boldsymbol{e}_i$ within $\boldsymbol{e}_{\mathrm{single}}$ exhibit entangled semantics~\cite{entangle_1,entangle_2,entangle_3}: each per-image embedding contains spectral components aligned with the shared identity as well as components conflicting with it, as we analyze in \cref{sec:analysis}. Because prior single-prompt methods~\cite{1p1s,zigzag} rescale every spectral component uniformly, they amplify or suppress these identity-aligned and identity-conflicting components together. To address this and answer \emph{which components} to control, we propose a selective embedding modification approach, illustrated in \cref{fig:overall_method}(a), that \textit{amplifies semantic components shared by identity and expression prompts}, while \textit{suppressing identity-irrelevant components}.

Briefly, our method consists of three steps performed in both the expression and suppression stages:
(1) constructing a reference embedding and its reference vector, which captures the directions to amplify or suppress and anchors component selection;
(2) selecting components by their similarity to the reference vector;
(3) rescaling the selected components and reconstructing the embeddings.

\paragraph{Selective expression.}
First, we construct a reference embedding by concatenating identity and expression embeddings, $\boldsymbol{e}_{\mathrm{ref\text{-}exp}} = [\boldsymbol{e}_{\mathrm{id}}, \boldsymbol{e}_{\mathrm{exp}}]$.
The generated image must reflect the target expression while preserving the shared identity. Therefore, the amplification reference is constructed from both embeddings. We then apply Singular Value Decomposition (SVD) separately to $\boldsymbol{e}_{\mathrm{exp}}$ and $\boldsymbol{e}_{\mathrm{ref\text{-}exp}}$, obtaining right singular vectors and singular values
$\{\boldsymbol{v}_i, \sigma_i\}_{i=1}^{r_{\mathrm{exp}}}$ for $\boldsymbol{e}_{\mathrm{exp}}$ and
$\{\bar{\boldsymbol{v}}_j, \bar{\sigma}_j\}_{j=1}^{r_{\mathrm{ref\text{-}exp}}}$ for $\boldsymbol{e}_{\mathrm{ref\text{-}exp}}$,
where $r_{\ast} = \min(L_{\ast}, d)$ denotes the number of singular components and $L_{\mathrm{ref\text{-}exp}} = L_{\mathrm{id}} + L_{\mathrm{exp}}$.
We define a reference vector:
\begin{equation}
\boldsymbol{v}_{\mathrm{ref\text{-}exp}} = \sum_{j=1}^{r_{\mathrm{ref\text{-}exp}}} \bar{\sigma}_j \bar{\boldsymbol{v}}_j,
\label{eqn:reference_vector}
\end{equation}
which is obtained by weighting each right singular vector by its corresponding singular value. To resolve the sign ambiguity of SVD, we orient each singular vector to have a nonnegative inner product with the token-wise mean of $\boldsymbol{e}_{\mathrm{exp}}$~\cite{bro2008resolving}. Second, we compute cosine similarity between singular vectors $\boldsymbol{v}_i$ and $\boldsymbol{v}_{\mathrm{ref\text{-}exp}}$, defining an adaptive threshold:
\[
\zeta_{\mathrm{exp}} = \frac{1}{r_{\mathrm{exp}}} \sum_{i=1}^{r_{\mathrm{exp}}} \cos(\boldsymbol{v}_i,\boldsymbol{v}_{\mathrm{ref\text{-}exp}}).
\]
Components with similarity above $\zeta_{\mathrm{exp}}$ are marked as expression-relevant and will be amplified. Since $\zeta_{\mathrm{exp}}$ is derived from the similarities themselves, the selection requires no manually tuned threshold. Finally, we amplify the corresponding singular values as follows:
\begin{equation}
\begin{aligned}
\sigma_i &\leftarrow U_{\mathrm{exp}}(\sigma_i),
\quad \text{if} \quad \cos(\boldsymbol{v}_i,\boldsymbol{v}_{\mathrm{ref\text{-}exp}}) > \zeta_{\mathrm{exp}},
\end{aligned}
\label{eqn:express_process}
\end{equation}
with the amplification function defined as \(U_{\mathrm{exp}}(\sigma) = \beta e^{\alpha \sigma}\sigma\), where $\alpha$ and $\beta$ (and $\alpha'$, $\beta'$ below) are hyperparameters specified in \cref{app:implementation_detail}.
We then reconstruct the amplified text embeddings using the updated singular values.

\paragraph{Selective suppression.}
We use the identity embedding $\boldsymbol{e}_{\mathrm{id}}$ as the reference embedding $\boldsymbol{e}_{\mathrm{ref\text{-}sup}}$ to select identity-irrelevant components in suppression embeddings.
This preserves the components that support the shared identity while downscaling only those irrelevant to it, so that suppressing a per-image prompt does not weaken identity.
Similar to the selective expression step, we apply SVD separately to the suppression embedding $\boldsymbol{e}_{\mathrm{sup}}$ and the reference embedding $\boldsymbol{e}_{\mathrm{ref\text{-}sup}}$, obtaining singular vectors and singular values $\{\boldsymbol{v}_i, \sigma_i\}_{i=1}^{r_{\mathrm{sup}}}$ and $\{\bar{\boldsymbol{v}}_j, \bar{\sigma}_j\}_{j=1}^{r_{\mathrm{id}}}$, respectively.
We define the reference vector $\boldsymbol{v}_{\mathrm{ref\text{-}sup}}$ from $\boldsymbol{e}_{\mathrm{ref\text{-}sup}}$ in the same way as in \cref{eqn:reference_vector}. Next, we compute cosine similarity between each singular vector $\boldsymbol{v}_i$ of $\boldsymbol{e}_{\mathrm{sup}}$ and the reference vector $\boldsymbol{v}_{\mathrm{ref\text{-}sup}}$, and define the adaptive threshold
\[
\zeta_{\mathrm{sup}} = \frac{1}{r_{\mathrm{sup}}} \sum_{i=1}^{r_{\mathrm{sup}}} \cos(\boldsymbol{v}_i,\boldsymbol{v}_{\mathrm{ref\text{-}sup}}).
\]
Components with similarity below $\zeta_{\mathrm{sup}}$ are regarded as identity-irrelevant and will be suppressed. Finally, we downscale the singular values of the selected components using
\begin{equation}
\begin{aligned}
\sigma_i &\leftarrow D_{\mathrm{sup}}(\sigma_i),
\quad \text{if} \quad \cos(\boldsymbol{v}_i,\boldsymbol{v}_{\mathrm{ref\text{-}sup}}) < \zeta_{\mathrm{sup}},
\end{aligned}
\label{eqn:suppression_process}
\end{equation}
where the downscaling function is defined as \(D_{\mathrm{sup}}(\sigma)=\beta' e^{-\alpha' \sigma}\sigma\).
We apply selective suppression iteratively to all $\boldsymbol{e}_{\mathrm{sup}_j}$, $j \in \{1,\ldots,k\}\setminus\{i\}$, and then reconstruct the suppressed text embeddings using the updated singular values.

The exponential forms of $U_{\mathrm{exp}}$ and $D_{\mathrm{sup}}$ follow the singular-value rescaling schedule standard in the single-prompt setting~\cite{1p1s,zigzag}, but unlike its uniform application to every singular value, we rescale only the components selected by the reference-guided criteria above, as analyzed in \cref{sec:analysis}.
We set $\alpha_{\mathrm{exp}}=0.025$ and $\alpha_{\mathrm{sup}}=-0.01$; \cref{app:sensitivity} shows our method is robust to both.

\subsection{Semantic Projection onto Padding Embeddings (PAD)} \label{sec:semantic_container}
Answering \emph{where to write} additional semantics, we leverage padding embeddings as semantic containers to enhance prompt alignment.
Prior work rescales EOT embeddings uniformly~\cite{1p1s}, suppresses content through the EOT embeddings of a single CLIP encoder~\cite{get}, or analyzes padding tokens without controlling them~\cite{paddingtone}. However, none identifies \emph{which} padding embeddings can carry semantics in multi-encoder backbones, nor actively \emph{writes} target semantics into them.
Our analysis in \cref{sec:pad} shows that in SD3.5, only OpenCLIP padding embeddings retain substantial prompt-related semantics, unlike CLIP and T5 embeddings that predominantly encode prompt-irrelevant (``dummy'') semantics. We therefore utilize only OpenCLIP padding embeddings as containers.
FLUX employs a single text encoder, excluding the pooled-embedding encoder. In this single-encoder regime, even predominantly dummy padding embeddings still influence image generation and can serve as semantic containers, as shown in \cref{app:padding_semantic}.
We write per-image semantics into the containers in two steps. First, we incorporate semantics aligned with the expression embedding:
\begin{equation} \boldsymbol{e}_{\mathrm{pad}} \leftarrow (1-\gamma)\, \boldsymbol{e}_{\mathrm{pad}} + \gamma\, \boldsymbol{e}_{\mathrm{proj\text{-}exp}}. \end{equation}
Then, we remove redundancy aligned with the suppression embedding:
\begin{equation} \boldsymbol{e}_{\mathrm{pad}} \leftarrow \boldsymbol{e}_{\mathrm{pad}} - \gamma\, \boldsymbol{e}_{\mathrm{proj\text{-}sup}}. \end{equation}
Here, $\gamma$ is a blending weight, and $\boldsymbol{e}_{\mathrm{proj\text{-}exp}}$ and $\boldsymbol{e}_{\mathrm{proj\text{-}sup}}$ are the projections of $\boldsymbol{e}_{\mathrm{pad}}$ onto the row spaces of the expression and suppression embeddings, respectively:
\begin{equation}
\boldsymbol{e}_{\mathrm{proj\text{-}c}} = \boldsymbol{e}_{\mathrm{pad}}\, P_{\mathrm{c}},
\qquad
P_{\mathrm{c}} = \boldsymbol{e}_{\mathrm{c}}^\top (\boldsymbol{e}_{\mathrm{c}} \boldsymbol{e}_{\mathrm{c}}^\top)^{\dagger} \boldsymbol{e}_{\mathrm{c}},
\end{equation}
where $(\cdot)^{\dagger}$ denotes the pseudo-inverse and $\mathrm{c} \in \{\mathrm{exp}, \mathrm{sup}\}$; for $\boldsymbol{e}_{\mathrm{proj\text{-}sup}}$, we set $\boldsymbol{e}_{\mathrm{sup}}$ to the concatenation of all suppression embeddings $[\boldsymbol{e}_{\mathrm{sup}_1},\ldots,\boldsymbol{e}_{\mathrm{sup}_{k-1}}]$. Both projections are computed from the padding embeddings before the update, and the two steps are applied sequentially.

The refined padding embeddings are then integrated into selective expression: we concatenate only the first $L_{\mathrm{exp}}$ of them, denoted $\boldsymbol{e}_{\mathrm{pad\text{-}sub}}\in\mathbb{R}^{L_{\mathrm{exp}}\times d}$, with $\boldsymbol{e}_{\mathrm{exp}}$, reinforcing expression-specific semantics.
Using only a subset prevents prompt-irrelevant padding semantics from interfering with the amplification of expression-related components, as shown in \cref{fig:analysis}(b5).
We set the blend weight to $\gamma=0.85$; our method is also robust to this choice, as shown in \cref{app:sensitivity}.

\begin{figure}[t]
    \centering
    \includegraphics[width=\linewidth]{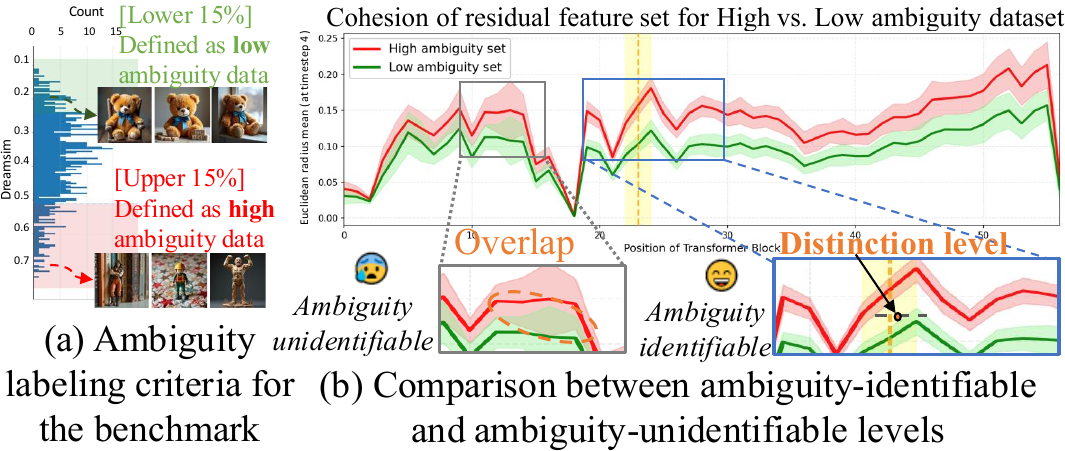}
    \caption{
    Criterion for labeling identity prompt ambiguity.
    (a) Benchmark examples of high- and low-ambiguity prompts.
    (b) FLUX residual-feature cohesion across transformer blocks: at ambiguity-identifiable blocks (blue box) the high- and low-ambiguity sets separate clearly, while elsewhere they overlap.
    }
    \label{fig:ambiguity}
\end{figure}

\subsection{Adaptive Feature Sharing (AFS)} \label{sec:ambiguity}
When the identity description is detailed, embedding modification alone preserves identity consistency; highly ambiguous descriptions (e.g., ``a man''), however, require an additional identity anchor.
Existing methods apply image feature sharing uniformly regardless of this distinction~\cite{consistory,characonsist,storydiffusion}, needlessly constraining diversity for already-specific prompts.
Thus, we propose \textit{Adaptive Feature Sharing}, which automatically assesses identity ambiguity and applies feature sharing selectively, answering \emph{when to constrain}.

We measure the ambiguity of an identity prompt from the spread of its per-scene features: from a transformer block, we take the feature for each pair $\{p_{\mathrm{id}},p_{i}\}$ and compute their mean Euclidean radius, the mean distance from the centroid.
A large radius means the character's appearance varies from scene to scene rather than staying consistent, which signals that the prompt does not pin down a single appearance; we label such a prompt \textit{highly ambiguous} when its radius exceeds a calibrated threshold.
To calibrate it, we build a held-out prompt benchmark disjoint from our evaluation data. It contains high- and low-ambiguity prompt sets, shown in \cref{fig:ambiguity}(a) and detailed in \cref{app:ambiguity}.
On this benchmark, we select the block and timestep whose radius best separates the two sets, shown in \cref{fig:ambiguity}(b): block 23 and timestep 4 of 28 for FLUX, and block 25 and timestep 3 for SD3.5.
Since both timesteps fall early in the 28-step schedule, detection requires only a few denoising steps and adds little overhead.
Calibrated \emph{once}, this criterion generalizes across subjects and styles (96.7\% accuracy, AUC 0.990) and remains effective on real images (0.81 accuracy on Co3D).

We supply this anchor only for prompts classified as \textit{highly ambiguous}. As illustrated in \cref{fig:overall_method}(b), we first cache image features conditioned solely on $p_{\mathrm{id}}$, then replace the corresponding features during generation. This forces all scenes to share a common appearance. The shared feature type follows validated recipes: residual features in FLUX~\cite{reflex} and key-value features in SD3.5~\cite{exploringmmdit}.
The result is an architecture-agnostic identity anchor engaged only on demand.

\section{Experiments}
\begin{figure*}[t!]
    \centering
    \includegraphics[width=\linewidth]{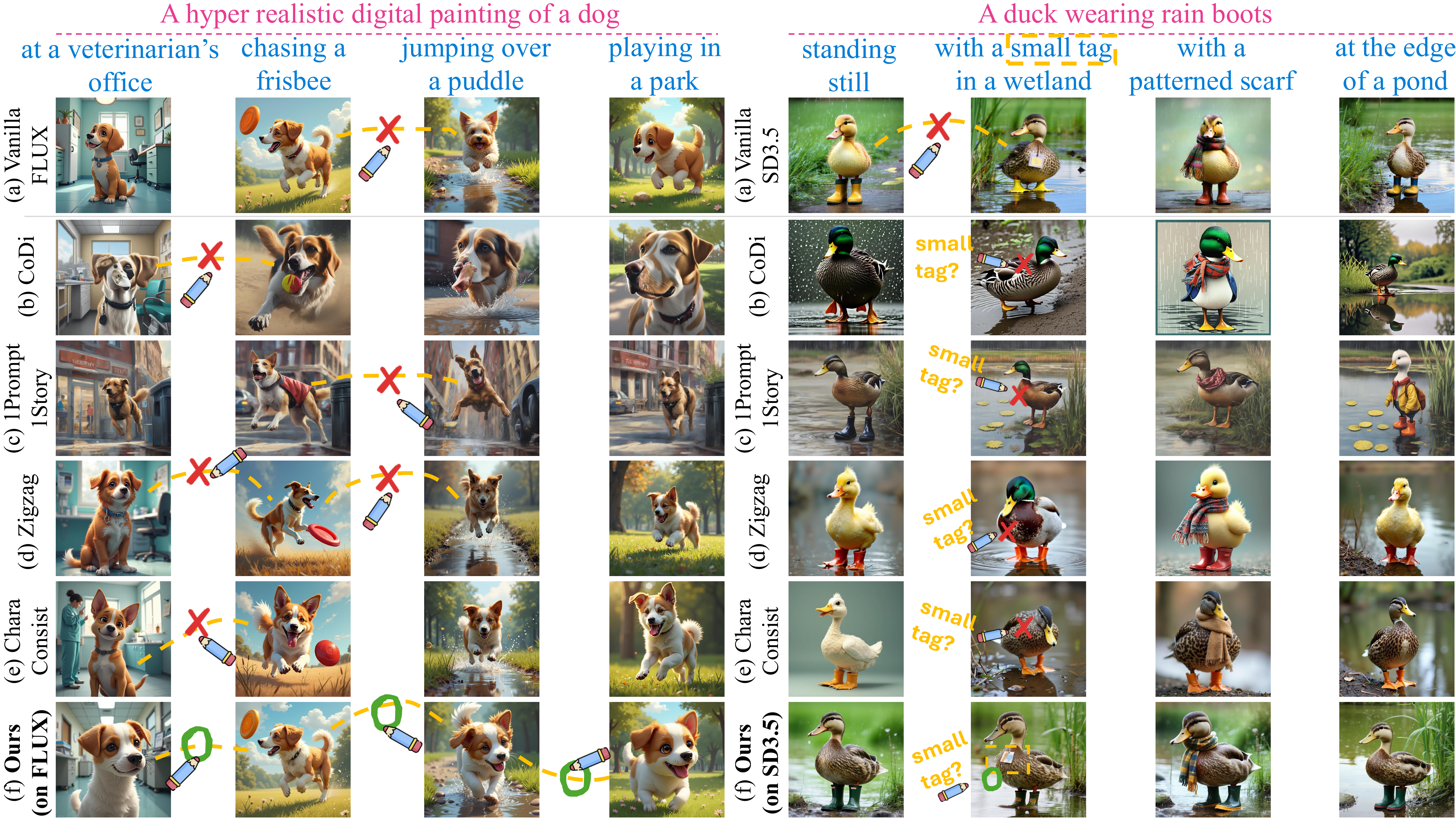}
    \caption{Qualitative comparison. The identity prompts are pink, and per-image prompts are blue. Our method best maintains identity and prompt alignment; others struggle with text details (e.g., missing tag) or identity consistency.}
    \label{fig:qual}
\end{figure*}

\subsection{Balanced Evaluation: SeeSaw}\label{sec:seesaw}
Existing protocols report identity consistency and per-image prompt alignment separately, typically using DreamSim~\cite{dreamsim} and VQAScore~\cite{vqascore}. This separation does not explicitly measure the trade-off between the two objectives. The prompt-alignment metric itself is also limited because VQA-based scores are right-skewed and tend to saturate~\cite{eval_evaluators}. When the identity prompt is long, images that ignore per-image attributes (e.g., pose or background) can still score high.
\textit{SeeSaw} addresses both issues: it balances identity consistency and per-image prompt alignment while explicitly testing whether the generated image faithfully reflects the per-image semantics.

Formally, given an identity prompt $p_{\mathrm{id}}$ and per-image prompts $\{p_1,\dots,p_k\}$, let $x_i$ denote the image generated for $[p_{\mathrm{id}}, p_i]$ in multi-prompt settings~\cite{codi,characonsist,consistory}, or for $p_i$ as the \emph{expression prompt} in single-prompt settings~\cite{1p1s,zigzag,decorstory}. We measure identity consistency using the DreamSim distance $D(x_i,x_j)$ between images sharing $p_{\mathrm{id}}$, and prompt alignment using the VQAScore
$t_i=\mathrm{VQA}(x_i,[p_{\mathrm{id}},p_i])$. We define $d_i$ as
\begin{equation}
d_i = 1 - \frac{1}{k-1} \sum_{\substack{j=1 \\ j \neq i}}^{k} D(x_i, x_j).
\end{equation}
The $(1-\cdot)$ converts distance to similarity, and min-max scaling aligns its range with the VQAScores.

To account for prompt representation imbalance, we compute the gap
$\Delta_i=a_i-t_i$
using the per-image prompt alignment score
$a_i=\mathrm{VQA}(x_i,p_i)$.
A positive $\Delta_i$ indicates that the per-image prompt is well represented, whereas a negative value indicates that it is under-represented. We then aggregate identity consistency and prompt alignment after correcting for prompt representation imbalance:
\begin{equation}
h_i=
\frac{
2\left(
d_i
-\mu\max(-\Delta_i,0)
+\tau\max(\Delta_i,0)
\right)t_i
}{
\left(
d_i
-\mu\max(-\Delta_i,0)
+\tau\max(\Delta_i,0)
\right)
+t_i+\varepsilon
},
\end{equation}
where $\varepsilon>0$ ensures numerical stability. The terms
$\mu\max(-\Delta_i,0)$ and
$\tau\max(\Delta_i,0)$
control the penalty and reward strengths. We set $\mu=\tau=0.5$, assigning equal weights to the penalty and reward terms.

\textit{SeeSaw} complements standard metrics by jointly reporting per-image alignment and identity consistency. Additional analyses on prompt imbalance using MSR-VTT videos, comparisons with standard evaluation metrics, and sensitivity to $\mu,\tau$ are provided in \cref{app:cqs_weight}.

\subsection{Experimental Setup}

As our method is training-free, we compare against the vanilla diffusion baselines~\cite{sdxl,flux,sd3}, which directly condition on concatenated identity and per-image prompts, and other recent training-free identity-consistent generation methods, including U-Net-based (CoDi~\cite{codi}, 1Prompt1Story~\cite{1p1s}) and DiT-based (CharaConsist~\cite{characonsist}, Zigzag~\cite{zigzag}) approaches.
The evaluation benchmark and comparison details are provided in \cref{app:experimental_setting}.

\subsection{Comparison}

\begin{table}[t]
  \centering
  \resizebox{\columnwidth}{!}{%
    \begin{tabular}{lccc}
    \toprule
    \textbf{Model} & \textbf{\textit{SeeSaw}} $\uparrow$ & \makecell{Per-image text \\ alignment $\uparrow$} & \makecell{Identity \\ consistency $\downarrow$} \\
    \midrule
    Vanilla SDXL & 0.481 & 0.619 & 0.400 \\
    1Prompt1Story & 0.391 & 0.442 & \underline{0.281} \\
    CoDi & 0.488 & 0.585 & 0.304 \\
    \midrule
    Vanilla FLUX & 0.584 & 0.723 & 0.347 \\
    CharaConsist & 0.519 & 0.637  & 0.368 \\
    Zigzag & 0.661 & 0.796 & 0.309 \\
    \textbf{Ours (FLUX)} & \underline{0.664} & 0.786 & \underline{0.281} \\
    \midrule
    Vanilla SD3.5 & 0.636 & \textbf{0.820} & 0.367 \\
    \textbf{Ours (SD3.5)} & \textbf{0.681} & \underline{0.809} & \textbf{0.272} \\
    \bottomrule
    \end{tabular}
  }
  \caption{Quantitative comparison with training-free identity-consistent generation methods. We highlight the best score in \textbf{bold} and the second-best with \underline{underline}. Our method achieves state-of-the-art performance on the \textit{SeeSaw}, effectively balancing text alignment and identity consistency.}
  \label{tab:main_results}
\end{table}

\Cref{tab:main_results} reports \textit{SeeSaw} alongside the two standard criteria: per-image text alignment and identity consistency.
Our method attains the highest \textit{SeeSaw} on both backbones, indicating that it better maintains the balance between identity consistency and per-image text alignment.
On the SD3.5 backbone, ours improves \textit{SeeSaw} over the vanilla model, with markedly better identity consistency while keeping alignment near vanilla.
Among FLUX-based methods, ours attains the best \textit{SeeSaw} while achieving markedly better identity consistency than Zigzag, 0.281 vs.\ 0.309. It is also over $7\times$ faster, taking 3.0 to 3.6 minutes per five-image set versus Zigzag's 26.1, as reported in \cref{tab:latency}.
1Prompt1Story attains high identity consistency but substantially worse per-image alignment, consistent with our analysis in \cref{sec:analysis} that uniform rescaling suppresses per-image semantics together with identity-irrelevant ones.
As shown in \cref{fig:qual}, CharaConsist fails to preserve identity under highly ambiguous prompts such as ``dog'' or ``duck'' despite feature sharing, and, like CoDi, is sensitive to the quality of estimated masks and transport.
Zigzag improves alignment through its sampling strategy but favors it over identity, rather than achieving a balanced trade-off.
Overall, combining selective embedding modification with adaptive feature sharing balances per-image alignment and identity consistency and remains robust even for ambiguous identity prompts.
Additionally, \cref{app:commercial} provides further examples and qualitative comparisons with commercial models.

\subsection{Ablation Study}
\begin{table}[t]
  \centering
  \resizebox{0.95\columnwidth}{!}{%
    \begin{tabular}{lccc}
    \toprule
     & \textbf{\textit{SeeSaw}} $\uparrow$ & \makecell{Per-image text \\ alignment $\uparrow$} & \makecell{Identity \\ consistency $\downarrow$} \\
    \midrule
    (i) SD3.5 & \underline{0.636} & \textbf{0.820} & 0.367 \\
    (ii) SD3.5 + STM & 0.610 & 0.777 & 0.357 \\
    (iii) SD3.5 + STM\&PAD & 0.620 & 0.791 & 0.364 \\
    (iv) SD3.5 + AFS & 0.606 & 0.772 & \underline{0.320} \\
    (v) \textbf{Ours} & \textbf{0.681} & \underline{0.809} & \textbf{0.272} \\
    \bottomrule
    \end{tabular}
  }
  \caption{Ablation study of our components on SD3.5. %
  }
  \label{tab:ablation}
\end{table}

\Cref{tab:ablation} ablates each component on SD3.5.
STM ((i) vs.\ (ii)) improves identity consistency (0.367$\rightarrow$0.357) without modifying identity embeddings, confirming that identity semantics are partially encoded across per-image embeddings.
PAD ((ii) vs.\ (iii)) restores prompt alignment (0.777$\rightarrow$0.791) at a marginal identity cost, validating padding embeddings as semantic containers.
AFS ((i) vs.\ (iv)) strengthens identity consistency (0.367$\rightarrow$0.320) at some alignment cost.
Combining all three (v) yields the best \textit{SeeSaw} (0.681): identity consistency improves by 0.095 over the baseline while alignment stays within 0.011 of the vanilla backbone.
This synergy is expected: residual features encode only coarse identity information~\cite{reflex, exploringmmdit}, so our enriched text and padding embeddings guide feature sharing without harming per-image fidelity.
On SD3.5, whose vanilla alignment is exceptionally strong, the full combination is required, as \textit{SeeSaw} rewards balanced gains on both criteria.
The FLUX ablation, qualitative results, and additional ablations are provided in \cref{app:ablation}.

\subsection{Analysis}\label{sec:analysis}
\begin{figure}[t]
    \centering
    \includegraphics[width=\linewidth]{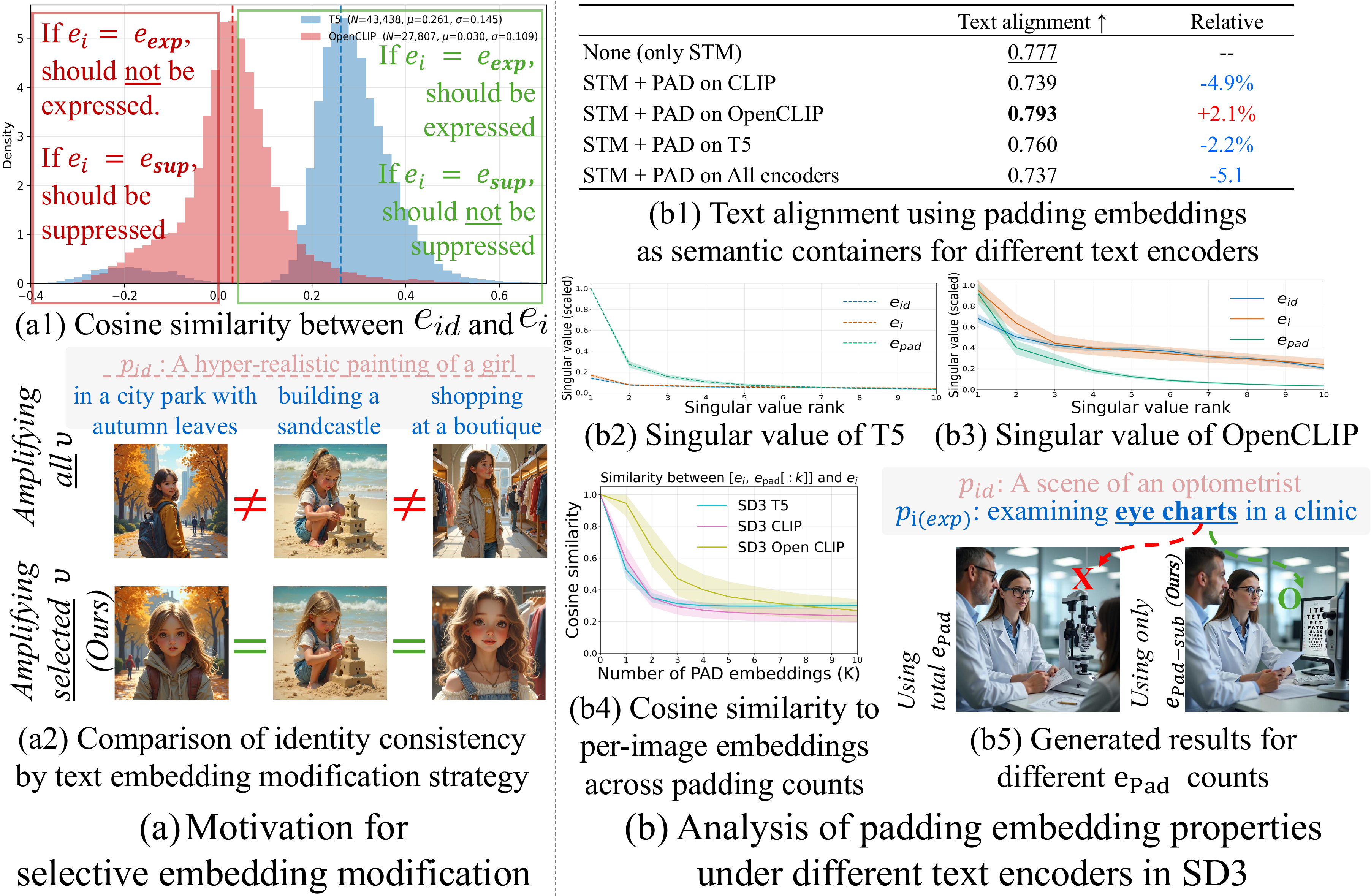}
    \caption{Analysis and validation of our text embedding modification. 
    We use about 500 samples from the benchmark for this analysis. 
     } 
    \label{fig:analysis}
\end{figure}

\paragraph{Motivation for selective embedding modification.}
\Cref{fig:analysis}(a1) shows that per-image embeddings are not orthogonal to the identity embedding: each contains components positively aligned with $\boldsymbol{e}_{\mathrm{id}}$ (green box), carrying semantics shared by the description and the identity, as well as negatively aligned components (red box) that conflict with it.
As defined in \cref{sec:svd}, selective expression amplifies only the former, improving expression fidelity and identity coherence together.
To isolate the effect of the selection rule, we compare against a \emph{uniform} variant that differs in exactly one factor: the same exponential schedule is applied to \emph{all} components, as in prior single-prompt methods~\cite{1p1s,zigzag}.
\Cref{fig:analysis}(a2) contrasts the two variants: uniform rescaling (upper) injects conflicting components and erases identity-supporting ones, causing semantic drift and reduced identity consistency, whereas our selective rescaling (lower) amplifies only the aligned components and suppresses only the identity-irrelevant ones, preserving identity coherence without sacrificing per-image fidelity.
\Cref{app:text_embedding_plot} shows that this entanglement appears in every text encoder of FLUX and SD3.5, so we apply selective expression and suppression to all encoder embeddings.

\paragraph{Semantic containers in padding space.} \label{sec:pad}
In \cref{fig:analysis}(b), we analyze the padding embeddings of SD3.5's multiple text encoders.
\Cref{fig:analysis}(b1) shows that semantic projection improves text alignment only for OpenCLIP.
\Cref{fig:analysis}(b2) shows why: T5 padding embeddings are dominated by a single large singular value, a prompt-irrelevant (``dummy'') direction.
\Cref{fig:analysis}(b4) confirms this, as adding padding tokens rapidly decreases cosine similarity to the original per-image embeddings.
SD3.5's CLIP EOT embeddings and FLUX's T5 embeddings behave similarly (demonstrated in \cref{app:padding_semantic}).
\Cref{fig:analysis}(b4, b5) further show that, even for OpenCLIP, projecting onto all padding tokens eventually degrades alignment whereas restricting to a subset preserves it; this motivates the subset selection in \cref{sec:semantic_container}.
We attribute this to encoder count: with multiple encoders, SD3.5 can ignore dummy padding directions, whereas FLUX relies on a single T5 encoder, allowing even dummy padding semantics to influence generation~\cite{paddingtone}.

Beyond consistent generation, our selective expression and padding-based semantic containers also extend to text-guided image editing; \cref{app:extension} shows that applying them to FLUX Kontext improves text alignment across edit-prompt types.

\section{Conclusion}
We traced the identity-alignment trade-off in consistent generation to component-level semantic entanglement in text embeddings, a structural cause that the prompt-level uniform rescaling of prior single-prompt methods cannot resolve.
AsemConsist resolves it by controlling individual spectral components, answering three questions prior methods leave open: \emph{which} components to rescale, via reference-guided selection; \emph{where} to write semantics, via padding embeddings that act as containers in multi-encoder backbones; and \emph{when} to constrain images, via ambiguity-adaptive feature sharing engaged only when needed.
We further proposed \textit{SeeSaw}, a metric that jointly measures identity consistency, per-image alignment, and their balance, while exposing the pitfall of alignment scores inflated by long identity prompts.
AsemConsist achieves the best identity-alignment balance on both FLUX and SD3.5.

\clearpage
\bibliography{sec/references}
\clearpage
\appendix

\renewcommand{\thetable}{A\arabic{table}}
\renewcommand{\thefigure}{A\arabic{figure}}
\setcounter{figure}{0}
\setcounter{table}{0}

\begin{figure*}[!t]
    \centering

    \begin{subfigure}[t]{0.44\linewidth}
        \centering
        \includegraphics[width=\linewidth]{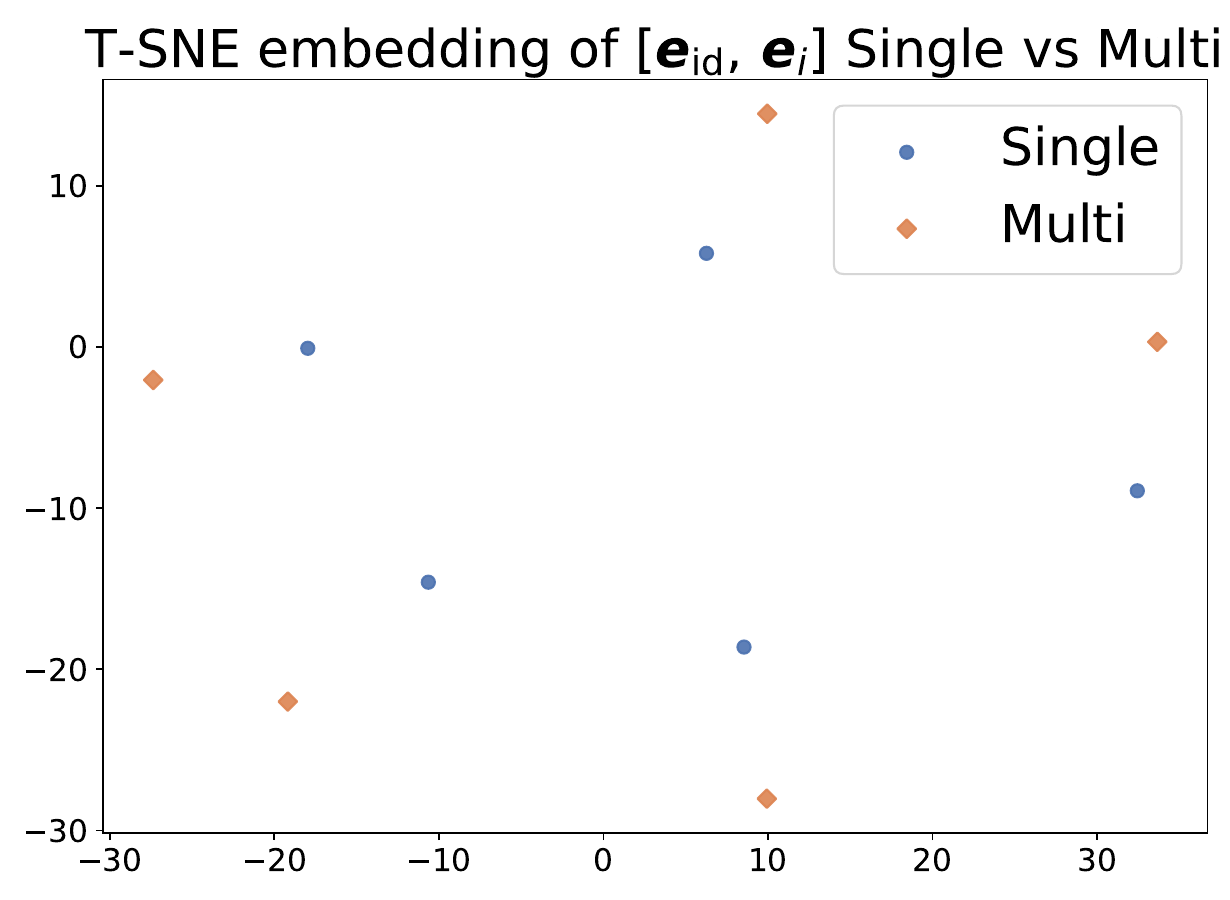}
        \caption{t-SNE comparison of single vs multi prompt embeddings.}
        \label{fig:tsne_1}
    \end{subfigure}
    \hfill
    \begin{subfigure}[t]{0.49\linewidth}
        \centering
        \includegraphics[width=\linewidth]{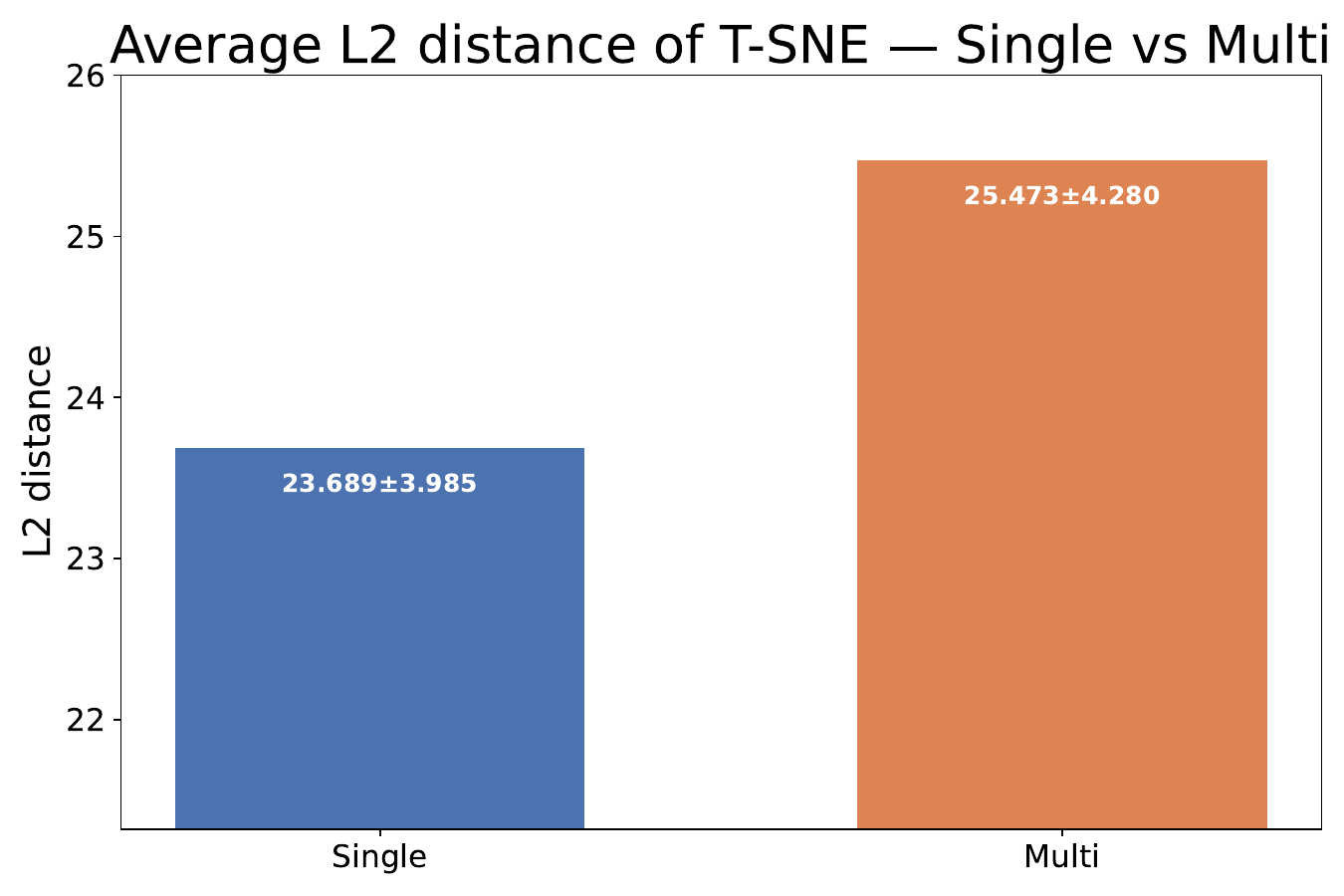}
        \caption{Average pairwise $L_2$ distances of single vs multi.}
        \label{fig:tsne_2}
    \end{subfigure}
    \caption{Single-prompt encoding improves identity semantic cohesion in the T5 text encoder. Compared with the multi-prompt setting, single-prompt embeddings cluster more tightly in t-SNE space and show smaller average pairwise $L_2$ distances.}
    \label{fig:tsne}
\end{figure*}
\begin{figure}[t]
    \centering
    \includegraphics[width=\linewidth]{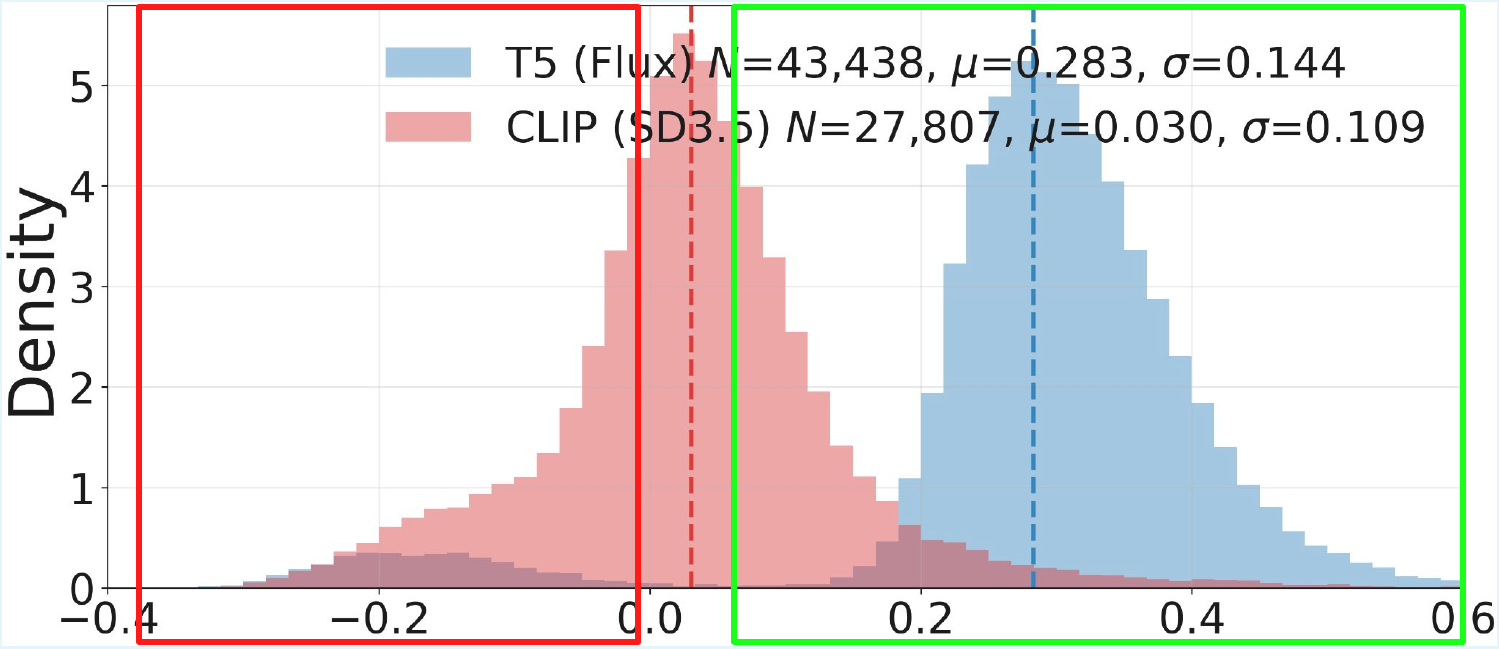}
    \caption{Cosine similarity between $e_{id}$ and $e_{i}$} 
    \label{fig:swr_sim}
\end{figure}

\begin{figure*}[t]
    \centering
    \includegraphics[width=\textwidth]{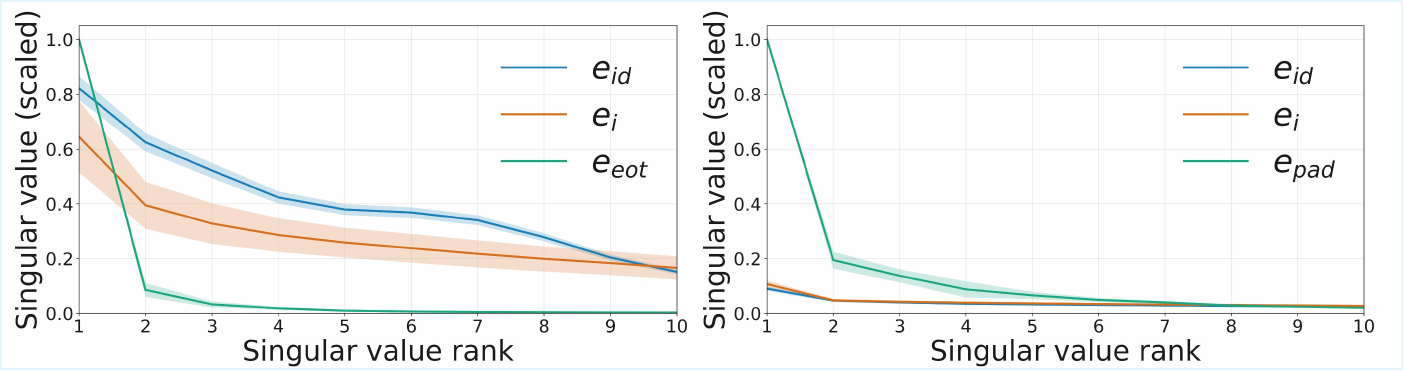}
    \caption{Singular value spectra of the EOT and padding embeddings: \\
    (\textbf{a}) CLIP in SD3.5 (\textbf{left}) and (\textbf{b}) T5 in FLUX (\textbf{right}).}
    \label{fig:sv_dominance}
\end{figure*}

\section*{Supplementary Materials}

\section{Text Embedding Analysis for Identity-Consistent Generation}

\subsection{Single Prompt Improves Identity Consistency in T5}\label{app:single_prompt}

In this section, we demonstrate the effectiveness of using a single combined prompt (identity prompt and all per-image prompts at once) compared to multiple separate prompts (identity prompt paired individually with each per-image prompt) in preserving identity consistency. Specifically, we analyze embedding cohesion in the T5 text encoder~\cite{t5} used by FLUX~\cite{flux} and SD3.5~\cite{sd3}, analogous to the analysis conducted on SDXL's CLIP embeddings~\cite{sdxl, clip} in 1Prompt1Story~\cite{1p1s}.

As introduced in \cref{sec:problem_setting}, we construct a single combined prompt by concatenating the identity and per-image prompts:
\[
p_{\mathrm{single}} = [p_{\mathrm{id}},\, p_1,\, \ldots,\, p_k].
\]
The T5 encoder $E_{\mathrm{T5}}(\cdot)$ generates the corresponding embeddings:
\[
\boldsymbol{e}_{\mathrm{single}} = [\boldsymbol{e}_{\mathrm{id}},\, \boldsymbol{e}_1,\, \ldots,\, \boldsymbol{e}_k] \in \mathbb{R}^{L_{\mathrm{single}} \times d},
\]
where $d$ denotes the embedding dimension and each $\boldsymbol{e}_\ast \in \mathbb{R}^{L_{\mathrm{\ast}} \times d}$ corresponds to $p_\ast$. We extract identity and per-image embeddings by splitting this combined embedding based on token indices, yielding pairs $[\boldsymbol{e}_{\mathrm{id}},\, \boldsymbol{e}_i]$.

Conversely, in the multi-prompt setting, the identity prompt is concatenated individually with each per-image prompt, encoded separately as:
\[
[\boldsymbol{e}^{(i)}_{\mathrm{id}},\, \boldsymbol{e}^{(i)}_{\text{multi}}] = E_{\mathrm{T5}}([p_{\mathrm{id}},\, p_i]),
\quad i\in \{1,\ldots,k\}.
\]

For evaluation, we use the official ConsiStory benchmark~\cite{consistory}, consisting of 122 prompts and 727 identity/per-image pairs. We project the embeddings from both single- and multi-prompt settings onto a shared 2D space using t-SNE. We then calculate the average pairwise $L_2$ distance between projected embeddings:
\[
d_{\mathrm{avg}}^{\text{t-SNE}} 
= \frac{1}{k(k-1)} \sum_{i \neq j} \|\mathbf{z}_i - \mathbf{z}_j\|_2,
\]
where $\mathbf{z}_i \in \mathbb{R}^2$ represents the t-SNE projection of embedding $i$.

As shown in \cref{fig:tsne_1}, embeddings from the single-prompt setting form significantly tighter clusters than those from the multi-prompt setting. \cref{fig:tsne_2} quantitatively confirms that single-prompt embeddings consistently have smaller average $L_{2}$ distances, indicating stronger semantic cohesion.

This result implies that the single-prompt setting effectively encourages the T5 encoder to maintain consistent identity semantics across diverse scenes, whereas the multi-prompt approach encodes identities independently per image, resulting in more dispersed representations. Consequently, our single-prompt formulation promotes stronger semantic coherence and improved identity consistency in the embedding space.

\subsection{Identity-Related Semantic Structure in CLIP (SD3.5) and T5(FLUX)}\label{app:text_embedding_plot}

In this section, we provide additional experimental results supporting that the proposed selective expression and selective suppression strategies can be applied across different text encoder environments. 

To further verify this observation, we perform the same analysis on the T5 text encoder used in FLUX and the CLIP encoder used in SD3.5. As shown in \cref{fig:swr_sim}, both encoders exhibit components of $e_i$ that are positively correlated with $e_{id}$ (green box) as well as components that are negatively correlated with it (red box). These results show that the semantic entanglement structure, in which identity-related and identity-irrelevant components coexist within per-image embeddings $e_i$, consistently appears across text encoders used in different diffusion backbones.

\section{Implementation Detail} \label{app:implementation_detail}

\subsection{Backbone and GPU setting}
We implement our method on top of the \texttt{FLUX.1-dev} backbone for the FLUX experiments and \texttt{stable-diffusion-3.5-large} for the SD3.5 experiments. All experiments are conducted on a single NVIDIA RTX A6000 GPU.

\subsection{Adaptive Feature Sharing}
For the SD3.5 backbone~\cite{sd3}, Adaptive Feature Sharing (\textbf{AFS}) is applied to image key-value features from the initial 30\% of transformer blocks at denoising steps 3 to 20 (of 28), previously identified as appearance-rich feature levels~\cite{exploringmmdit}.
For the FLUX backbone~\cite{flux}, \textbf{AFS} utilizes the residual stream~\cite{reflex} at transformer blocks \([0, 1, 2, 17, 18]\), known as semantically influential "vital layers"~\cite{stableflow}, during early denoising steps \(t \in [1, 6]\).
Importantly, these selections of layers and steps are not hyperparameters tuned specifically in our study; rather, they rely solely on previously established feature levels known to encode appearance information.

\subsection{Selective Text Embedding Modification and Padding Projection}

STM is applied with the following scaling parameters:
\[
\alpha_{\text{exp}} = 0.025,\quad
\beta_{\text{exp}} = 1.0,\quad
\alpha_{\text{sup}} = -0.01,\quad
\beta_{\text{sup}} = 0.05,
\]
and PAD is applied with a blend weight of $\gamma = 0.85$.

\subsection{Experiments settings}
We report only the settings shared across all experiments. Experiment-specific details, including benchmarks, subsets, runs, seeds, and reported variation, are provided in the corresponding sections.

\paragraph{Random Seeds}
Randomness arises only from the initial latent noise. Unless otherwise specified, all experiments use a fixed seed of 42. To measure sampling variation, we additionally evaluate the default configuration with seeds $\{0, 7, 42\}$ and report the resulting SeeSaw variation.

\section{Discriminative Criterion for Highly Ambiguous Descriptions} \label{app:ambiguity}

As described in \cref{sec:ambiguity}, we define the ambiguity level of an identity description. 
A \textit{high-ambiguity} description produces diverse appearances when images are generated from an identity prompt $p_{\mathrm{id}}$ combined with multiple per-image prompts $(p_{1},\dots,p_{k})$. 
In contrast, a \textit{low-ambiguity} description results in consistent character appearances across the prompt set.

To identify transformer blocks and timesteps where ambiguity can be clearly distinguished, we first construct high- and low-ambiguity sample sets. 
Using a large language model (ChatGPT), we generated approximately 1000 prompt sets of the form $\{p_{\mathrm{id}}, p_{1}, \dots, p_{k}\}$, where each identity prompt is paired with six per-image prompts ($k=6$), yielding 6000 per-image prompts (7000 prompts in total). 
The prompts ranged from simple descriptions (e.g., ``a zebra'') to highly detailed ones (e.g., ``a 16-year-old girl with wavy chestnut hair, a slender frame, and soft brown eyes''). 
Subjects included animals, people, objects, food, and other categories, with no overlap with our evaluation dataset.

We synthesized images from these prompts using the base diffusion models (SD3.5 and FLUX) and measured visual similarity within each prompt set using DreamSim. 
The lowest 15\% similarity scores were labeled as the high-ambiguity set, while the highest 15\% were labeled as the low-ambiguity set.

Next, we computed the mean Euclidean radius of residual features from each transformer block at every denoising timestep for each prompt set. 
A higher Euclidean radius indicates more dispersed samples (high ambiguity), whereas a lower value indicates tightly clustered samples (low ambiguity).

For FLUX, the clearest separation occurred at transformer block 23 and timestep 4 (out of 28 steps):

\begin{itemize}
    \item High ambiguity: $0.1571 \pm 0.0135$
    \item Low ambiguity: $0.1032 \pm 0.0116$
\end{itemize}

The effect size (Cohen's $d$) is 4.26, indicating minimal overlap between the distributions. 
Based on this observation, we set the ambiguity threshold at a Euclidean radius of 0.1285:

\begin{itemize}
    \item High ambiguity if mean Euclidean radius $>$ 0.1285
    \item Low ambiguity if mean Euclidean radius $\leq$ 0.1285
\end{itemize}

For SD3.5, the most discriminative location is transformer block 25 and timestep 3 (out of 28 steps):

\begin{itemize}
    \item High ambiguity: $0.1295 \pm 0.0103$
    \item Low ambiguity: $0.0965 \pm 0.0077$
\end{itemize}

The effect size (Cohen's $d$) is 3.62, again indicating minimal overlap. 
The resulting ambiguity threshold is:

\begin{itemize}
    \item High ambiguity if mean Euclidean radius $>$ 0.1126
    \item Low ambiguity if mean Euclidean radius $\leq$ 0.1126
\end{itemize}

We further evaluate robustness across styles and subjects. 
For style robustness, we evaluated style robustness by fixing the subject and per-image prompts (100 sets) and varying only the artistic style across three variants: \textit{pop art}, \textit{sketch}, and \textit{3D polygon mesh}.
Across all three style variants, our method consistently identified the same ambiguity-critical step–layer pairs, achieving 96.67\% accuracy and an AUC of 0.990.

For subject robustness, we construct a benchmark using Co3D~\cite{co3d} consisting of same-category but different-identity pairs (100 high-ambiguity cases). 
Using Gemini-generated captions and RF-Inversion~\cite{rfinversion}, we verify ambiguity detection and achieve 0.81 accuracy even on real images.

\section{Padding Embedding Analysis for the SD3.5 CLIP Encoder}\label{app:padding_semantic}

In this section, we analyze the CLIP encoder in SD3.5. Our goal is to further verify that semantic containers in the padding space are primarily observed in the OpenCLIP~\cite{openclip} encoder, while CLIP~\cite{clip} and T5~\cite{t5} embeddings are dominated by a prompt-irrelevant (``dummy'') direction and therefore exhibit relatively limited semantic representational capacity. 

To further support this observation, we analyze the singular value spectrum of the EOT embedding in the SD3.5 CLIP encoder. As shown in \cref{fig:sv_dominance}(a), the EOT embedding also exhibits a singular value distribution dominated by one large singular value, similar to the behavior observed in T5. This indicates that most of the variance in the embedding space is aligned with a single semantic direction. This result suggests that the EOT embedding of CLIP is also dominated by a prompt-irrelevant (``dummy'') semantic direction.

\subsection{Why Does Padding Projection Work for the T5 Encoder in FLUX?}
In this section, we analyze why padding projection works effectively in FLUX. Interestingly, padding projection produces a meaningful effect in FLUX even though the padding embeddings exhibit strong singular value dominance, as shown in \cref{fig:sv_dominance}(b). We interpret this behavior as arising from the structure of the text encoder used in FLUX. Unlike SD3.5, which performs conditioning with multiple text encoders, FLUX relies on a single text encoder (T5)~\cite{t5}. In models that employ multiple text encoders, even if padding embeddings converge to a prompt-irrelevant (``dummy'') direction, embeddings from other encoders can still provide sufficient semantic information, limiting the influence of padding embeddings. In contrast, because FLUX uses a single encoder, even padding embeddings that are close to the dummy direction can still exert a certain level of influence during the generation process. To further verify this interpretation, we conduct an ablation study in \cref{app:ablation}.

\begin{table*}[!t]
\centering
\small
\resizebox{0.95\linewidth}{!}{%
\begin{tabular}{l ccc c ccc}
\toprule
 & \multicolumn{3}{c}{\textbf{SDXL-based Models}} & & \multicolumn{3}{c}{\textbf{SD3.5-based Models}} \\
\cmidrule(lr){2-4} \cmidrule(lr){6-8}
\textbf{Metric} & \makecell{Vanilla \\ SDXL} & \makecell{1Prompt \\ 1Story} & CoDi & & \makecell{Vanilla \\ SD3.5} & \makecell{AsemConsist on SD3.5 \\ (Low ambiguity)$^{\dagger}$} & \makecell{AsemConsist on SD3.5 \\ (High ambiguity)$^{\dagger}$} \\
\midrule
Latency (minute) & 1.29 & 2.05 & 2.69 & & 1.96 & 7.37 & 9.01 \\
\bottomrule
\end{tabular}%
}

\resizebox{\linewidth}{!}{%
\begin{tabular}{lccccc}
\toprule
 & \multicolumn{5}{c}{\textbf{FLUX-based Models}} \\
\cmidrule(lr){2-6}
\textbf{Metric} & \makecell{Vanilla \\ FLUX} & \makecell{Chara \\ Consist} & Zigzag$^{\dagger}$ & \makecell{AsemConsist on FLUX \\ (Low ambiguity)} & \makecell{AsemConsist on FLUX \\ (High ambiguity)} \\
\midrule
Latency (minute) & 2.37 & 6.15 & 26.10 & 3.02 & 3.55 \\
\bottomrule
\end{tabular}%
}
\caption{Latency comparison across various generative models. The models are grouped by their base architectures (SDXL, SD3.5, and FLUX). $^{\dagger}$Measured with CPU offload.}
\label{tab:latency}
\end{table*}
\section{Comparison Details} \label{app:experimental_setting}

\subsection{Evaluation Benchmark}

For quantitative evaluation, we propose a new text benchmark. Existing benchmarks, such as ConsiStory+~\cite{1p1s}, use overly detailed identity prompts and relatively short per-image prompts. Repetitive style descriptions across samples often enable direct inference of character identity from prompts, complicating accurate assessments of identity consistency and per-image text alignment.

To address these issues, we constructed a new benchmark with assistance from a large language model (ChatGPT), guided by the following criteria:
\begin{itemize}
\item Subjects span eight categories: animal, human, fantasy and fairytales, toy, vehicles, plants, technology, and architecture. Each category includes between 10 and 20 recommended subjects.
\item Identity prompts are concise and general, facilitating diverse interpretations (\eg, "a cat").
\item Per-image text prompts incorporate varied poses, backgrounds, and additional attributes suitable for each subject.
\item Style descriptions comprise roughly 20\% of the prompt and emphasize general styles, such as "cartoon" or "sketch."
\end{itemize}

In total, we generated 90 subjects for evaluation.

\subsection{Competitors}

We compare our method against several training-free models for consistent identity generation. Below, we detail each competitor's evaluation setup and notable limitations. All competitors were evaluated using their official repositories.

\paragraph{FLUX}
To determine the scope of identity-consistent generation achievable by the backbone itself, we include FLUX~\cite{flux}, our backbone model, and the current state-of-the-art text-to-image diffusion model. Specifically, we use the \texttt{FLUX-dev} implementation from diffusers.

\paragraph{1Prompt1Story}
1Prompt1Story~\cite{1p1s} achieves consistency by uniformly rescaling the singular values of expression and suppression embeddings (SVR) and by identity-preserving cross-attention (IPCA). Because the rescaling is applied uniformly to all components, per-image semantics are suppressed together with identity-irrelevant ones, and when the identity prompt is ambiguous (e.g., ``a man''), identity preservation fails. Moreover, IPCA is tied to U-Net cross-attention, and its reliance on the SDXL backbone further limits text alignment compared to recent DiT models.

\paragraph{CharaConsist}
CharaConsist~\cite{characonsist} ensures character identity by employing RoPE-based feature correspondence in FLUX. This baseline performs well primarily when identity descriptions are detailed enough for FLUX itself to maintain consistency. CharaConsist explicitly requires separate prompts: a Character prompt (appearance), an Environment prompt (background), and an Action prompt (pose). In our evaluation, we directly use our identity prompts as Character prompts and separate Action and Environment prompts from per-image prompts via ChatGPT.

\paragraph{CoDi}
CoDi~\cite{codi} utilizes optimal transport in the early denoising steps to map reference identity features onto target images in a pose-aware manner. Subsequently, in the later denoising steps, the model refines identity using a selective cross-image attention mechanism. While these methods ensure consistent subject generation across diverse scenes , artifacts or unnatural poses can occur when the transport process fails. We employed the official implementation with the SDXL~\cite{sdxl} backbone to generate the samples.

\paragraph{Zigzag}
Zigzag~\cite{zigzag} decomposes each denoising step into zig, zag, and generation sub-steps. It injects subject-specific visual tokens exclusively during the zig phase and adds noise with a null prompt during the zag phase to prevent textual overfitting. However, the model can struggle to maintain the balance between identity and text alignment, as it relies on these two distinct methods to handle potentially conflicting semantic information, and its three-phase sampling incurs a substantial latency overhead (\cref{tab:latency}). We utilized the official implementation with the FLUX.1-dev base model to generate our samples.

\subsection{Evaluation Protocol}

\paragraph{Per-image Text Alignment}
For per-image text alignment, we adopt VQAScore~\cite{vqascore} with the official code. Following the original protocol, we query the model with the question ``Does this image show \{ \}?'' and measure the probability that the answer is ``Yes'', which serves as the alignment score for each image. We use Qwen3-VL-8B-Instruct~\cite{qwen3vl} as a VQA model. 

\paragraph{Identity Consistency}
To measure identity consistency across generated images, we calculate perceptual similarity using DreamSim~\cite{dreamsim}. DreamSim effectively captures semantic content, allowing us to compute similarity scores between all pairs of generated images. The final identity consistency score is obtained by averaging these values across all sample pairs.

\subsection{Inference Latency Comparison}
In this section, we compare the inference times of all base models and competing methods. To ensure fair comparisons, we evaluated all methods using a single NVIDIA RTX A6000 GPU on the same server. We define inference time as the total duration required to generate a fixed set of five images with consistent identity. Detailed results are presented in \cref{tab:latency}.

Under the same FLUX baseline, AsemConsist achieves the highest SeeSaw score while maintaining relatively fast inference times. Specifically, AsemConsist with FLUX takes approximately 3.02 to 3.55 minutes per set, which is significantly faster than CharaConsist (approximately 6.15 minutes per set) and Zigzag with CPU offload (approximately 26.1 minutes per set).

However, AsemConsist on SD3.5 exhibits longer inference times (7.37 to 9.01 minutes per set) due to the high capacity features used in adaptive feature sharing, which require CPU offloading as previously identified~\cite{exploringmmdit}. Nevertheless, considering that AsemConsist on SD3.5 leads overall in SeeSaw scores and provides excellent text alignment, the slightly increased inference time, about 3 times faster than Zigzag and only around 2 to 3 minutes longer compared to CharaConsist, is an acceptable tradeoff.

\section{Ablation Study on FLUX}\label{app:ablation}
\begin{figure}[t]
    \centering
    \includegraphics[width=\linewidth]{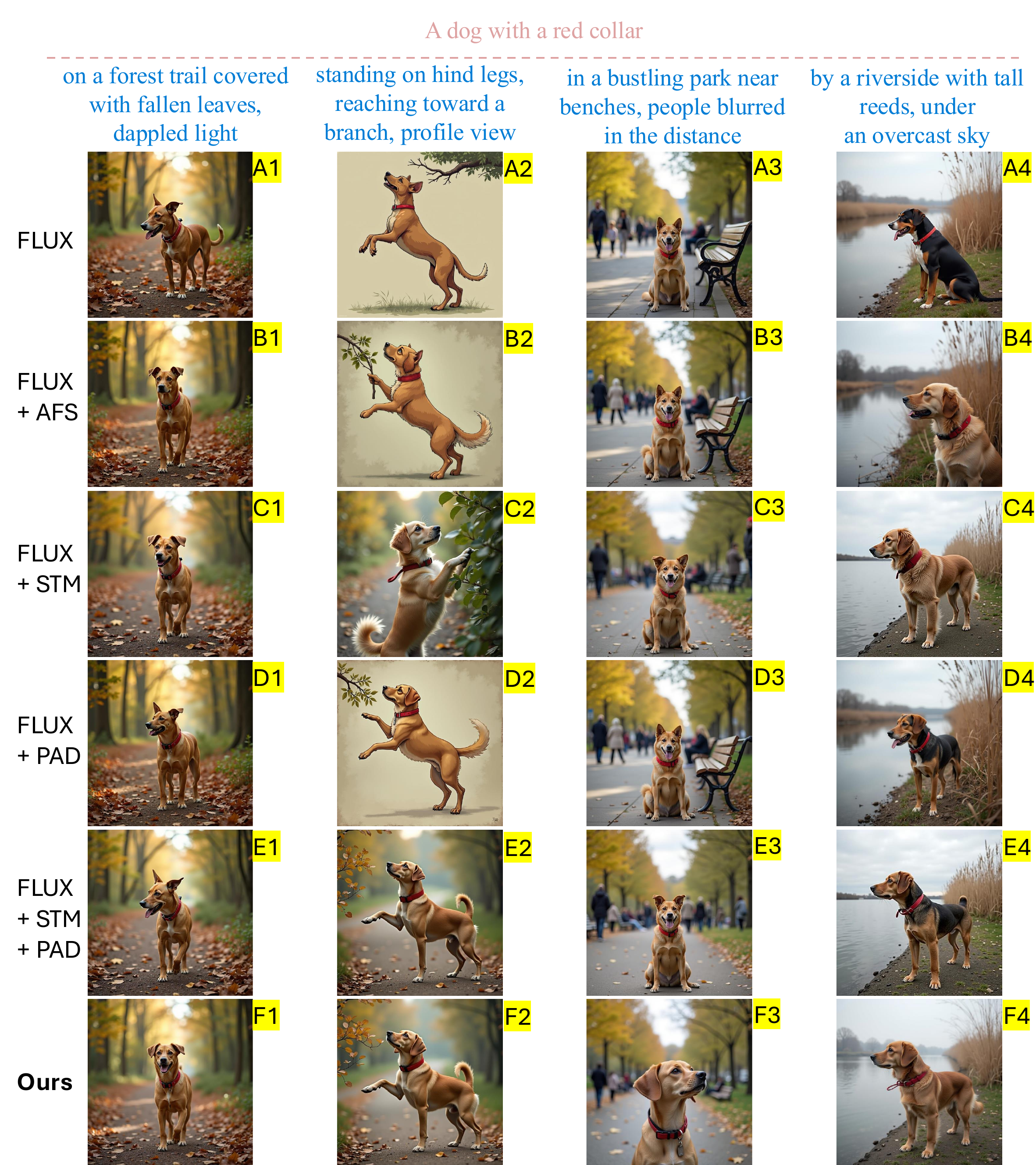}
    \caption{Qualitative results of ablation settings.}
    \label{fig:supp_abl}
\end{figure}
\begin{table}[!t]
  \centering
  \resizebox{\linewidth}{!}{%
    \begin{tabular}{lccc}
    \toprule
                    & \textbf{SeeSaw} $\uparrow$ &\makecell{Per-image text \\ alignment $\uparrow$} & \makecell{Identity \\ consistency $\downarrow$} \\
    \midrule
    (a) FLUX   & 0.488 & \textbf{0.631} & 0.429 \\
    (b) FLUX + STM & 0.541 & 0.625 & \underline{0.324} \\
    (c) FLUX + STM$\&$PAD & \underline{0.543} & \underline{0.628} & 0.326 \\
    (d) FLUX + AFS & 0.500 & 0.622 & 0.396 \\
    (e) \textbf{Ours} & \textbf{0.553} & 0.627 & \textbf{0.302} \\
    (f) FLUX + PAD & 0.491 & 0.619 & 0.408 \\
    \bottomrule
    \end{tabular}
  }
    \caption{Ablation study comparing the FLUX baseline, STM, STM+PAD, AFS, and our full model, with an additional projected PAD-only variant (FLUX+PAD).}

  \label{tab:ablation2}
\end{table}
In this section, we ablate each component on FLUX using a subset of 30 benchmark subjects and additionally examine the single-encoder interpretation of padding projection discussed in \cref{app:padding_semantic}. As shown in \cref{tab:ablation2}, the PAD-only setting (FLUX+PAD), where only padding projection is applied, improves identity consistency to a certain extent compared to the baseline FLUX model. This suggests that padding embeddings in FLUX can function as a basic semantic register for organizing per-image semantics.

However, in the PAD-only setting, the overall performance improvement remains limited, and text alignment shows a slight decrease. This suggests that padding embeddings alone behave more like a weak regularization mechanism and are not sufficient to fully organize the semantic structure.

In contrast, when STM and PAD are used together (STM+PAD), both identity consistency and text alignment improve simultaneously. This is because STM organizes the semantic structure related to expression, enabling padding embeddings to function effectively as semantic containers. These results show that the padding embedding space in FLUX is not a completely meaningless dummy space, but can serve as a potential semantic space that organizes per-image semantics through appropriate semantic projection. This effect is also observed in the qualitative comparison, where our model shows more consistent identities and more stable prompt alignment compared to the FLUX baseline. \cref{fig:supp_abl} illustrates the qualitative results of each ablation configuration:

\begin{enumerate}
    \item The FLUX baseline generates inconsistent identities across frames (See A1$\sim$A4).
    \item Adding AFS alone produces limited improvement in identity preservation(See B1$\sim$B3).
    \item Using only PAD improves identity consistency (A2 $\rightarrow$ D2, A4 $\rightarrow$ D4) to a limited degree but remains insufficient (D1$\sim$ D3).
    \item Using only STM also shows clear limitations in maintaining identity. (See C1$\rightarrow$ C3)
    \item Combining STM and PAD results in significantly better identity preservation.(D2 $\rightarrow$ E2, D4 $\rightarrow$ E4) 
    \item Adding AFS on top of STM+PAD (\textbf{Ours}) further improves identity consistency while maintaining strong text alignment. (See F1$\sim$F4)
\end{enumerate}

\section{Further Analysis of SeeSaw}
\label{app:cqs_weight}

\begin{table*}[t!]
\centering

\includegraphics[width=0.9\textwidth]{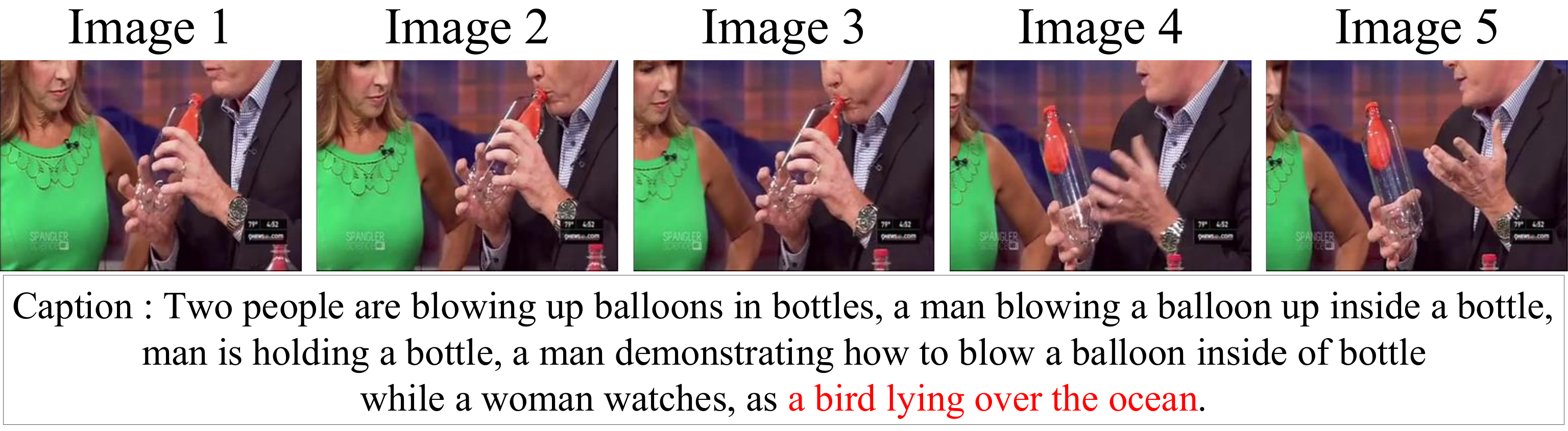}

\resizebox{\textwidth}{!}{%
\begin{tabular}{lccccc}
\toprule
 & Text alignment $\uparrow$ & DreamSim $\downarrow$ & $\Delta_i = a_i - t_i$ & Harmonic mean $\uparrow$ & SeeSaw $\uparrow$ \\
\midrule
Result & 0.866 (0.068) & 0.145 (0.158) & -0.810 (0.084) & 0.892 (0.055) & 0.664 (0.050) \\
\bottomrule
\end{tabular}
}

\caption{Comparison between the simple harmonic mean and SeeSaw under prompt imbalance. Values are reported as mean (std).}
\label{tab:hm_vs_seesaw}
\end{table*}

\subsection{Comparison with Seesaw and Harmonic Mean}

\begin{table}[t!]
\centering
\includegraphics[width=\linewidth]{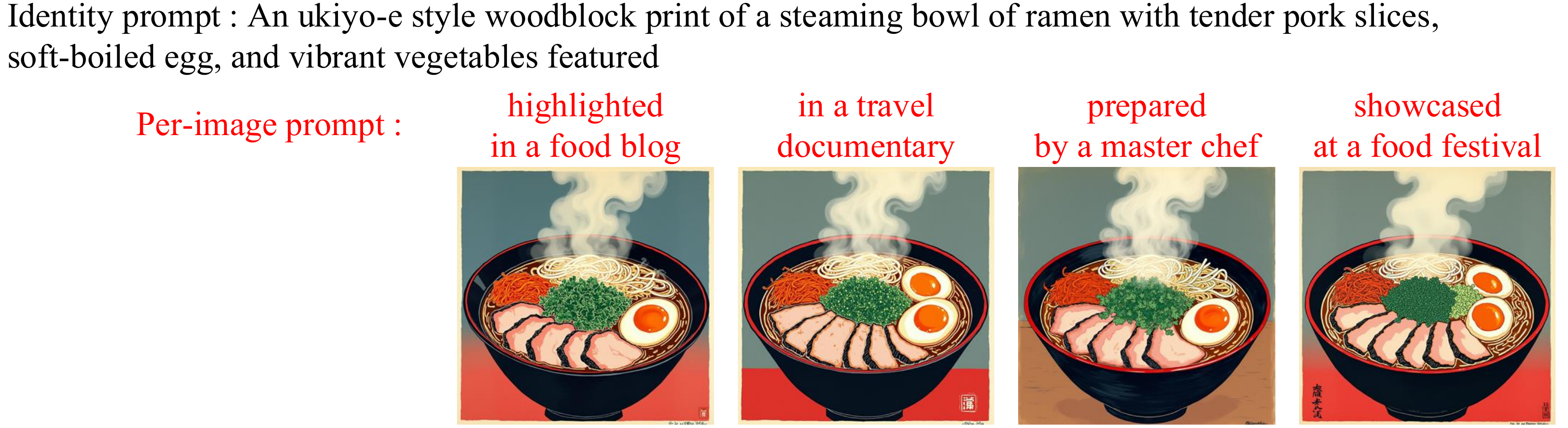}

\newcolumntype{Y}{>{\centering\arraybackslash}X}
\newcolumntype{L}{>{\raggedright\arraybackslash}p{3.3cm}}

\begin{tabularx}{\linewidth}{LYYYY}
\toprule
VQAScore total $\uparrow$ & 0.815 & 0.911 & 0.815 & 0.946 \\
VQAScore per-image $\uparrow$ & 0.400 & 0.351 & 0.454 & 0.369 \\
Dreamsim $\downarrow$ & 0.069 & 0.043 & 0.053 & 0.018 \\
Harmonic mean $\uparrow$ & 0.814 & 0.936 & 0.876 & 0.964 \\
\textbf{SeeSaw} $\uparrow$ & 0.667 & 0.567 & 0.691 & 0.582 \\
\bottomrule
\end{tabularx}

\caption{Comparison under a long identity prompt. 'VQAScore total' denotes the VQAScore derived from the combination of identity and per-image prompts. While the simple harmonic mean fails to account for per-image alignment, leading to an overinflated score, SeeSaw provides a robust evaluation that accurately captures alignment discrepancies at the individual image level.}
\label{tab:hm_vs_seesaw_long_id}
\end{table}

In this section, we further validate that SeeSaw is not merely the harmonic mean of identity consistency (DreamSim) and text alignment (VQAScore), but a metric that additionally captures representation imbalance between identity prompts and per-image prompts.

As shown in \cref{tab:hm_vs_seesaw_long_id}, which presents an example generated from one of the prompts in the ConsiStory+ benchmark~\cite{1p1s}, when the identity prompt is excessively long, the VQAScore can remain high even if the per-image text is not properly reflected. In such cases, as illustrated in the example, both the total VQAScore and DreamSim are measured as high even though not all per-image prompts are satisfied, which consequently leads to a high harmonic mean. In contrast, SeeSaw directly incorporates the discrepancy between the per-image VQAScore and the total VQAScore as a penalty, allowing the metric to reflect prompt imbalance and avoid overestimating performance when per-image semantics are not faithfully preserved.

To quantitatively analyze this phenomenon, we conduct experiments using the MSR-VTT~\cite{xu2016msr} real video dataset containing about 10,000 short web videos, each paired with multiple natural language captions (typically around 20). These captions describe the same visual scene using different expressions, making the dataset suitable for constructing prompt imbalance scenarios where correct captions dominate the prompt while an unrelated caption is appended.

From this dataset, we select 200 videos and extract five consecutive frames with the smallest DreamSim distance from each video, resulting in 1000 images in total. Since these frames originate from the same video, they maintain high visual similarity and thus strong identity consistency.

For each image, we construct a prompt using the ground-truth captions of the corresponding video and append one caption from a different video category to create an imbalanced prompt. We then compute the total prompt VQAScore as well as the per-image VQAScore corresponding to the appended caption.

For an explicit example, consider a video showing an experiment of inflating a balloon inside a bottle. The prompt may contain captions such as: \textit{two people are blowing up balloons in bottles, a man blowing a balloon up inside a bottle, a man is holding a bottle, a man demonstrating how to blow a balloon inside of a bottle while a woman watches,}
followed by an unrelated caption such as \textit{"a bird flying over the ocean".}

Table~\ref{tab:hm_vs_seesaw} summarizes the results. Because the sampled frames come from the same video, the DreamSim distance remains low (0.145), indicating strong identity consistency. In addition, the total prompt VQAScore is high (0.866) since most captions correctly describe the image. Consequently, the simple harmonic mean of the two scores is also high (0.892).

However, the appended caption is unrelated to the image, leading to a large discrepancy between the total prompt alignment and the per-image prompt alignment, with $a_i - t_i = -0.810$. Because SeeSaw explicitly penalizes this imbalance, the final SeeSaw score drops to 0.664.

These results demonstrate that the simple harmonic mean may overestimate performance when identity-related captions dominate the prompt. In contrast, SeeSaw captures the representation imbalance between identity and per-image prompts, providing a more faithful evaluation of identity-consistent generation.

\subsection{Effect of Weighting on SeeSaw Stability}\label{app:cqs_ablation}
\begin{table*}[!t]
    \centering

    \vspace{2mm}
    \resizebox{\textwidth}{!}{
    \begin{tabular}{cccccccccc}
    \toprule
    $\mu,\tau$ & \textbf{Ours(SD3.5)} & \underline{Ours(FLUX)} & ZigZag & SD3.5 & FLUX & CharaConsist & CoDi & SDXL & 1Prompt1Story \\
    \midrule
    0.1 & \textbf{0.6926} & \underline{0.6669} & 0.6639 & 0.6444 & 0.5859 & 0.5154 & 0.4979 & 0.4833 & 0.3903 \\
    0.2 & \textbf{0.6900} & \underline{0.6664} & 0.6635 & 0.6427 & 0.5857 & 0.5165 & 0.4957 & 0.4830 & 0.3907 \\
    0.3 & \textbf{0.6873} & \underline{0.6657} & 0.6631 & 0.6408 & 0.5853 & 0.5175 & 0.4933 & 0.4826 & 0.3910 \\
    0.4 & \textbf{0.6840} & \underline{0.6620} & 0.6610 & 0.6380 & 0.5840 & 0.5180 & 0.4900 & 0.4820 & 0.3910 \\
    0.5 & \textbf{0.6812} & \underline{0.6641} & 0.6611 & 0.6363 & 0.5842 & 0.5192 & 0.4884 & 0.4813 & 0.3911 \\
    0.6 & \textbf{0.6783} & \underline{0.6635} & 0.6611 & 0.6344 & 0.5837 & 0.5202 & 0.4854 & 0.4806 & 0.3915 \\
    0.7 & \textbf{0.6750} & \underline{0.6628} & 0.6603 & 0.6321 & 0.5829 & 0.5209 & 0.4824 & 0.4796 & 0.3915 \\
    0.8 & \textbf{0.6715} & \underline{0.6619} & 0.6593 & 0.6295 & 0.5820 & 0.5215 & 0.4792 & 0.4784 & 0.3913 \\
    0.9 & \textbf{0.6678} & \underline{0.6608} & 0.6583 & 0.6268 & 0.5810 & 0.5220 & 0.4759 & 0.4770 & 0.3911 \\
    1.0 & \textbf{0.6639} & \underline{0.6597} & 0.6571 & 0.6240 & 0.5798 & 0.5224 & 0.4723 & 0.4754 & 0.3907 \\
    \bottomrule
    \end{tabular}
    }
    \caption{
    SeeSaw comparison across different reward and penalty weights ($\mu,\tau$).
    \textbf{Ours (SD3.5)} consistently achieves the best performance, while \underline{Ours (FLUX)} remains the second-best across all settings.
    }
    \label{tab:cqs_weight_small}
    \vspace{0pt}
\end{table*}
This subsection analyzes the sensitivity of the proposed \textbf{SeeSaw} metric to variations in the reward and penalty weights $\mu$ and $\tau$. As shown in Tab.~\ref{tab:cqs_weight_small}, we jointly vary both weights from 0.1 to 1.0 to examine the stability of the metric across a wide range of configurations. Across all settings, \textbf{Ours (SD3.5)} consistently achieves the highest SeeSaw scores, while \underline{Ours (FLUX)} remains the second-best across all weight configurations. Other baselines such as ZigZag, SD3.5, and FLUX show lower scores throughout the entire range of weights.

As $\mu$ and $\tau$ increase, the influence of the balance penalty becomes stronger, which gradually decreases the scores of models that exhibit weaker trade-offs between identity consistency and text alignment. Nevertheless, the ranking of the top seven methods remains unchanged across all weight settings; the only change is a swap between CoDi and vanilla SDXL, which differ by less than 0.002, at $\mu=\tau\ge0.9$. In particular, the ranking among the competitive methods (Ours, Zigzag, and the vanilla backbones) is stable throughout, indicating that SeeSaw provides a stable and discriminative evaluation of identity-consistent generation even under large variations in the weighting parameters.

\subsection{Scope of SeeSaw}
We clarify the scope of our identity-consistent generation evaluation metric, \textit{SeeSaw}. To address the issue where per-image prompts might not be reflected in the generated images, \textit{SeeSaw} applies rewards and penalties based on the difference between scores reflecting both identity and per-image prompts, and scores reflecting only the per-image prompt.

We do not consider cases where only per-image prompts are reflected without the identity prompt. Such scenarios rarely occur in practice because identity prompts are positioned at the beginning of the prompt and are therefore typically incorporated during generation. To verify this, we measure the VQAScore using only the identity prompts (excluding style prompts) on the ConsiStory+ benchmark. The results show that samples with a VQAScore below 0.2 for the identity prompt account for only 67 out of 1100 samples on average across the SD3.5 and FLUX backbones, corresponding to approximately 6.1\% of the total samples.

\section{Extension to Text-Guided Image Editing}\label{app:extension}
\begin{figure*}[t]
    \centering
    \includegraphics[width=\linewidth]{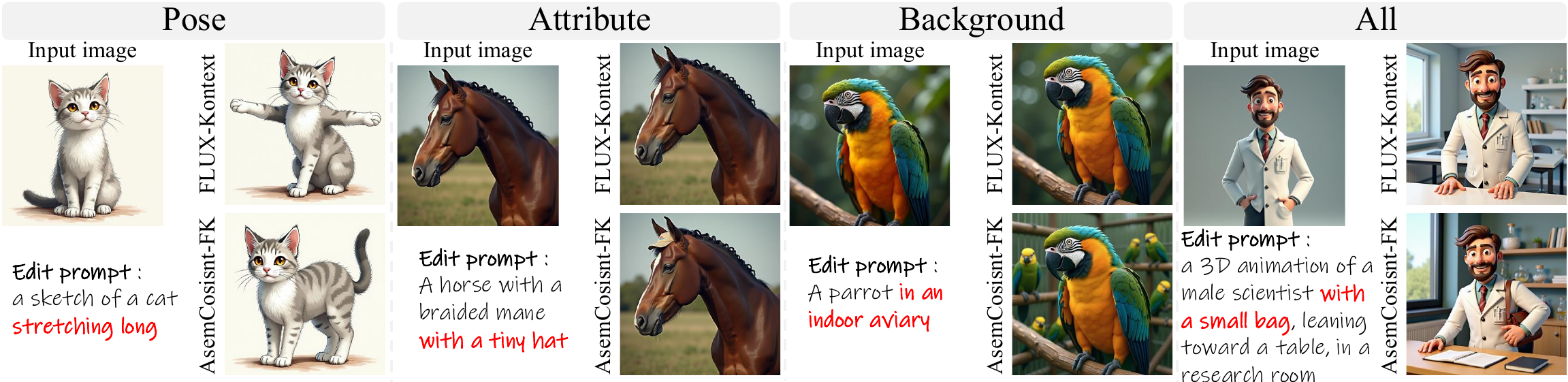}
    \vspace{0pt}

    \begingroup
    \renewcommand{\arraystretch}{0.93} %
    \setlength{\tabcolsep}{2.3pt}      %
    
    \small 
    \begin{tabular*}{\textwidth}{l @{\extracolsep{\fill}} cccccccc}
        \hline
         & \multicolumn{2}{c}{Pose} & \multicolumn{2}{c}{Attribute} & \multicolumn{2}{c}{Background} & \multicolumn{2}{c}{All} \\
        \cline{2-3} \cline{4-5} \cline{6-7} \cline{8-9}
        Metric & FK & AC-FK & FK & AC-FK & FK & AC-FK & FK & AC-FK \\
        \hline
        Text alignment & 0.410 & \textbf{0.452} & 0.665 & \textbf{0.681} & 0.639 & \textbf{0.657}  & 0.376 & \textbf{0.385} \\
        \hline
    \end{tabular*}
    \endgroup

    \caption{Comparison of AsemConsist with FLUX Kontext (AC-FK) and vanilla FLUX Kontext (FK) on four edit prompt variants: pose, attribute, background, and all (combining all three). Average VQAScores (text alignment) over 300 prompts per variant show AC-FK consistently outperforms FK.}
    \label{fig:category_results}
\end{figure*}

In this section, we provide additional analysis of how AsemConsist improves text-guided image editing models. We use FLUX Kontext~\cite{fluxkontext} as a representative example, where a reference image is edited according to a textual instruction. Unlike consistent character generation, image editing does not require preserving identity across multiple generated images. Therefore, only two components of AsemConsist are applicable in this setting: selective expression within selective text embedding modification and padding embeddings as semantic containers, while suppression prompts and ambiguity-adaptive feature sharing are unnecessary.

Selective expression preserves the identity of the reference image by enhancing only semantic components aligned with identity-related descriptions while suppressing unrelated modifications. As a result, the edited image better maintains the original appearance and avoids unrealistic changes, particularly when large pose variations are requested.

Padding embeddings serve as semantic containers for editing-specific semantics. By injecting semantic information into the padding space, they improve the model's ability to incorporate textual modifications that are otherwise difficult to express, including background changes and the introduction of new objects or attributes.

Consequently, integrating AsemConsist into FLUX Kontext substantially improves text alignment across diverse editing scenarios. As illustrated in \cref{fig:category_results}, our method alleviates several common failure cases of FLUX Kontext, including (1) failure to incorporate additional attributes described by text, (2) unrealistic pose editing, and (3) excessive reliance on the reference image background, which limits text-driven variations. These observations demonstrate that the proposed semantic manipulation strategies are not limited to consistent character generation but can also generalize effectively to text-guided image editing.

\section{Robustness to Scene Bias}\label{app:scenebias}
\begin{table}[t]
\centering

\includegraphics[width=0.85\linewidth]{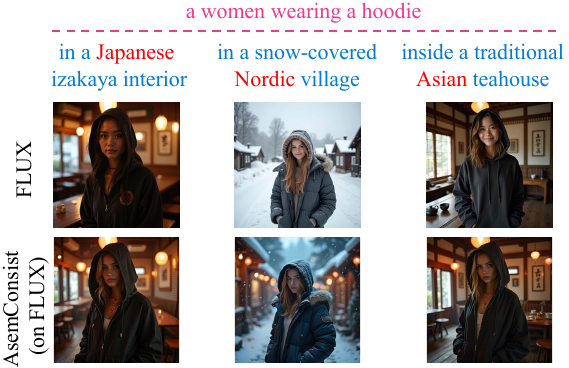}

\vspace{2mm}

\renewcommand{\arraystretch}{1.3}
\resizebox{0.6\linewidth}{!}{%
\begin{tabular}{lc}
\toprule
Model & DreamSim $\downarrow$ \\
\midrule
FLUX & 0.409 \\
\textbf{Ours FLUX} & \textbf{0.306} \\
\midrule
SD3.5 & 0.464 \\
\textbf{Ours SD3.5} & \textbf{0.342} \\
\bottomrule
\end{tabular}%
}

\caption{Qualitative results in various biased scenes (\textbf{a}) and DreamSim scores over 572 samples (\textbf{b}) for the backbone and our method.}
\label{fig:scene_bias}
\end{table}

In this section, we analyze the scene bias problem where identity can change when per-image prompts contain overly specific scene descriptions. 

To analyze this effect, we construct prompts composed of identity, per-image attributes, and scene descriptions. The scene prompts include environments with strong cultural or environmental cues, such as:

\begin{itemize}
\item \textit{in a snow-covered Nordic village}
\item \textit{inside a traditional Asian teahouse}
\item \textit{in a Moroccan bazaar}
\item \textit{in a Japanese izakaya interior}
\item \textit{in a European medieval town}
\end{itemize}

Such prompts can induce appearance priors associated with specific regions or cultures and may therefore lead to identity drift.

This phenomenon is also observed in existing backbone models. As shown in \cref{fig:scene_bias}(a), the vanilla FLUX model tends to generate subjects whose appearance reflects the Nordic environment when given prompts such as \textit{snow-covered Nordic village}, often failing to preserve the original identity.

In contrast, our method is more robust to scene bias. STM suppresses semantic components unrelated to identity while enhancing expression-related components, and AFS stabilizes identity-related features through residual feature sharing. Quantitative results further support this observation. As shown in \cref{fig:scene_bias}(b), DreamSim identity distance is measured over 572 samples:
FLUX (Ours / FLUX) = \textbf{0.306} / 0.409,
SD3.5 (Ours / SD3.5) = \textbf{0.342} / 0.464.
These results indicate that the proposed method more robustly preserves identity under scene-biased prompts.

\section{Limitations}

Our framework primarily maintains identity through text based adjustments and adaptive image feature sharing. However, it faces limitations in achieving high fidelity identity consistency when the identity prompt does not clearly specify the subject, such as natural scenes (e.g., "a lush mountain range") or ambiguous subjects like certain foods (e.g., spaghetti Bolognese). This limitation arises from our ambiguity aware distinctiveness metric, which is effective only when the subject is clearly identifiable.
Additionally, at the image feature level, such as FLUX's residual features or SD3.5's key and value features, the implicit appearance information is insufficient to distinctly define subjects lacking clear identity, including natural scenes or food items.
This limitation of handling ambiguous identity prompts aligns with findings in prior research~\cite{zigzag,consistory}.

Additionally, AsemConsist shows limited effectiveness when applied to the Qwen-Image backbone~\cite{qwen-image}. Our selective text embedding modification and padding-based semantic container modules rely on specific architectural properties: (i) the steerability of token-level identity semantics and (ii) the presence of unmasked padding tokens. These strategies do not directly generalize to backbones using large-scale vision-language model (VLM) encoders, such as Qwen-Image.
In such models, textual prompts are processed through deep LLM layers (Qwen2.5-VL), resulting in high-level semantic representations where identity signals are deeply integrated and regularized. This depth dilutes the effect of localized embedding modifications, making it difficult to isolate identity via SVD-based methods without disrupting global prompt adherence. Furthermore, Qwen-Image employs a strict attention masking strategy for padding tokens, rendering them non-participatory in the diffusion backbone's cross-attention patterns. Consequently, padding embeddings cannot effectively serve as semantic carriers in these architectures, limiting the applicability of our framework to VLM-conditioned diffusion models.

\section{Additional Qualitative Comparison Samples}
We provide additional qualitative comparisons in \texttt{Comparison.html}. Please double-click the HTML file to check the qualitative results.

\section{Comparison with Commercial Models}
\label{app:commercial}

\begin{figure*}[t]
\centering
\fbox{
\begin{minipage}{0.97\textwidth}
\small
\ttfamily

client.images.generate(\\
\hspace*{1.5em}model="gpt-image-2",\\
\hspace*{1.5em}prompt="""\\
\hspace*{3em}Task:\\
\hspace*{3em}Generate four images of the same character and arrange them\\
\hspace*{3em}into a single 2048x2048 image with a 2x2 grid layout.\\
\\
\hspace*{3em}Requirements:\\
\hspace*{3em}- Maintain exactly the same character identity.\\
\hspace*{3em}- Preserve <identity-defining characteristics>.\\
\hspace*{3em}- Apply only the modifications described in each\\
\hspace*{5em}image-specific prompt.\\
\hspace*{3em}- Follow every image-specific prompt faithfully.\\
\hspace*{3em}- Do not introduce unspecified appearance changes.\\
\\
\hspace*{3em}Identity Prompt:\\
\hspace*{3em}<Identity Prompt>\\
\\
\hspace*{3em}Image-specific Prompts:\\
\hspace*{3em}1. <Prompt 1>\\
\hspace*{3em}2. <Prompt 2>\\
\hspace*{3em}3. <Prompt 3>\\
\hspace*{3em}4. <Prompt 4>\\
\\
\hspace*{3em}Generate one 2048x2048 image containing the four results\\
\hspace*{3em}in a 2x2 grid while preserving the same identity across all\\
\hspace*{3em}four images.\\
\hspace*{1.5em}""",\\
\hspace*{1.5em}size="2048x2048",\\
\hspace*{1.5em}quality="medium",\\
\hspace*{1.5em}n=1,\\
)

\end{minipage}
}
\caption{
Prompt template and API call used for GPT Image 2 in our qualitative comparison.
}
\label{fig:commercial_prompt}
\end{figure*}
\begin{figure*}[t]
\centering

\includegraphics[
    width=\textwidth,
    trim={0 0 0 0},
    clip
]{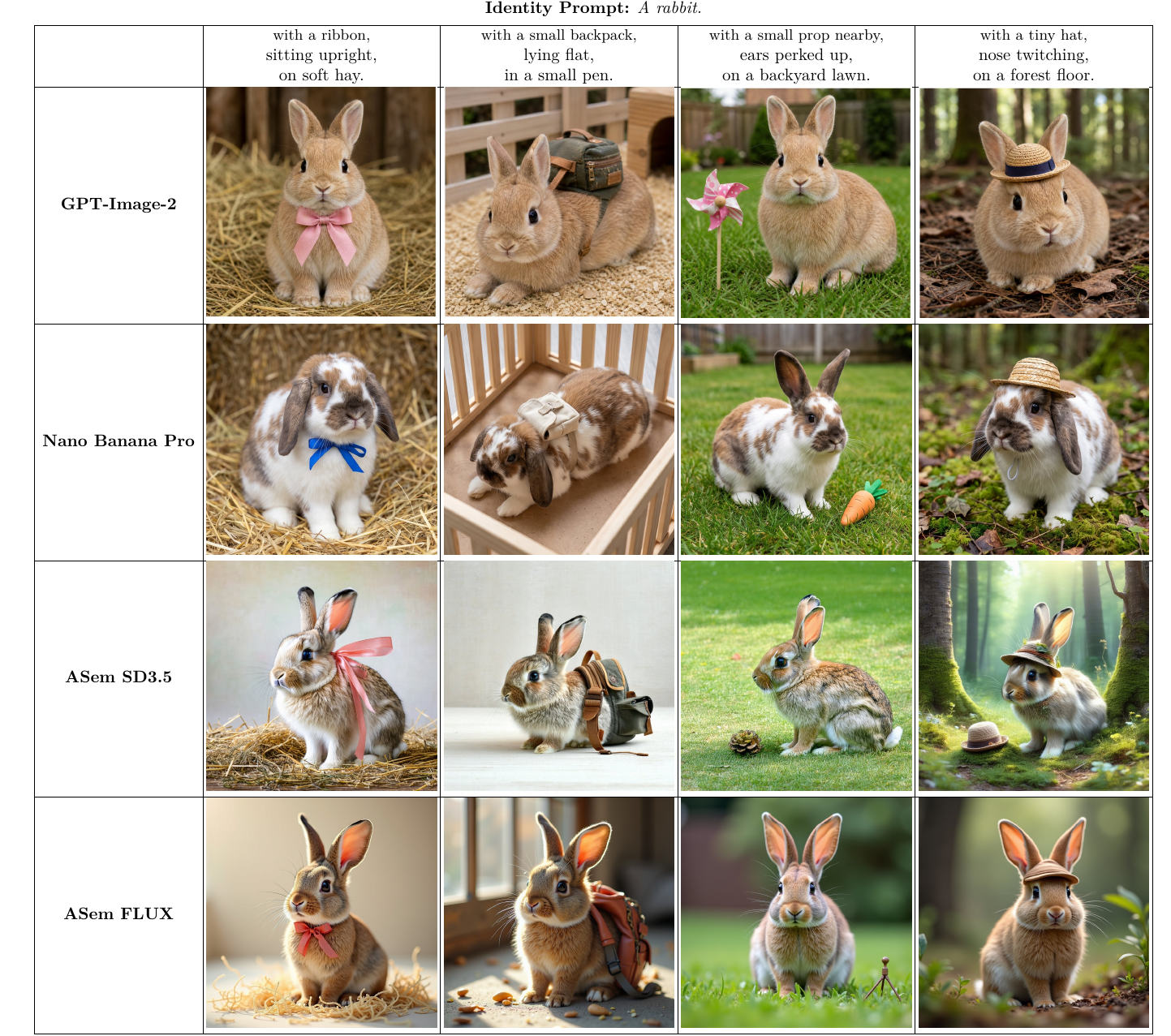}

\caption{Qualitative comparison with commercial image-generation models.}
\label{fig:commercial1}

\end{figure*}
\begin{figure*}[t]
\centering

\includegraphics[
    width=\textwidth,
    trim={0 0 0 0},
    clip
]{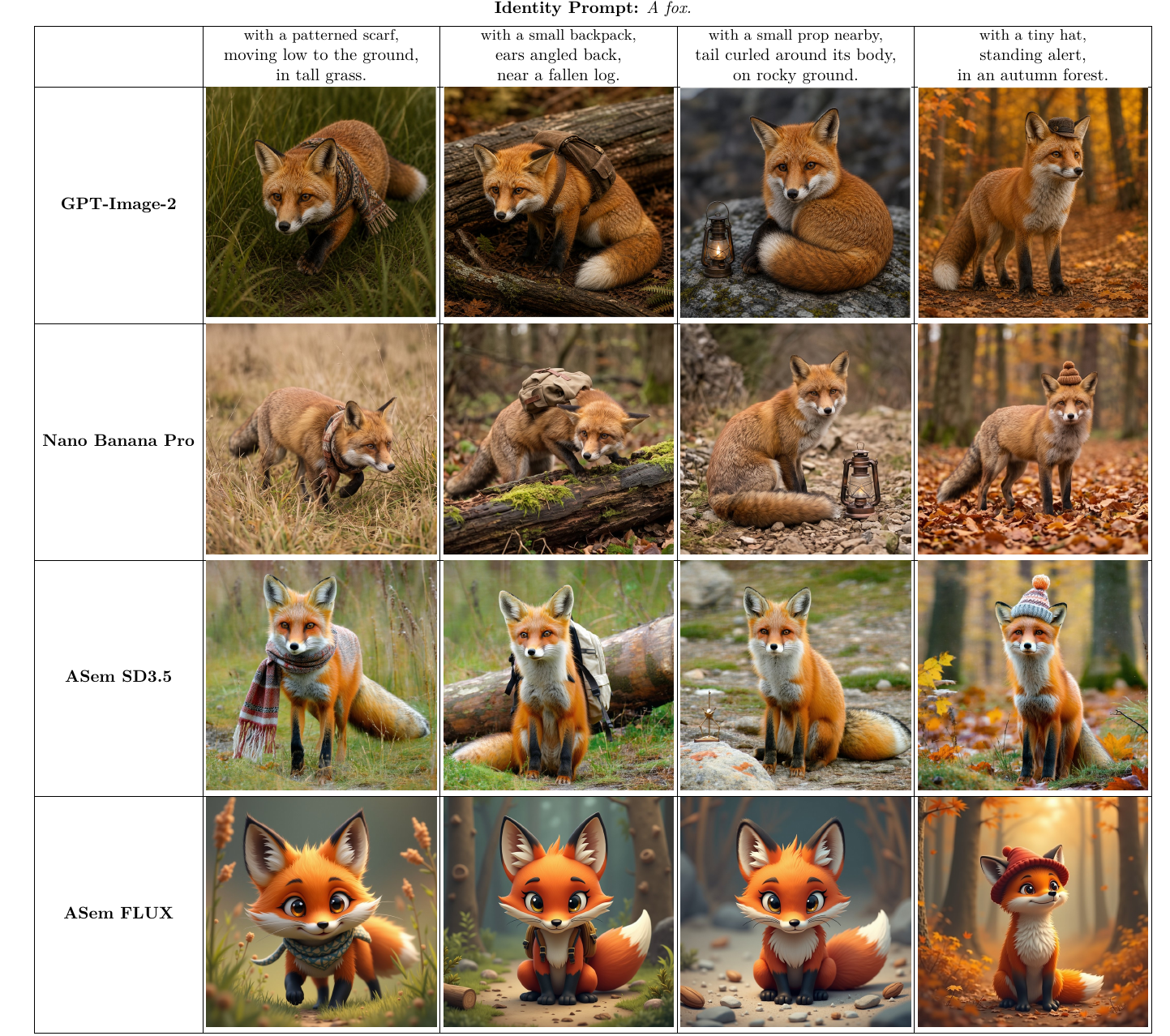}

\caption{Qualitative comparison with commercial image-generation models.}
\label{fig:commercial2}

\end{figure*}

We qualitatively compare AsemConsist with GPT-5.4 Pro and Nano Banana Pro on the consistent character generation task. All methods receive the same identity prompt and four image-specific prompts without any reference images or previously generated outputs. Each model is instructed to preserve the same character identity while faithfully following the image-specific prompts.

Each prompt set is generated as a single $2048\times2048$ image arranged in a $2\times2$ grid, with one panel corresponding to each image-specific prompt.

\subsection{Prompt Template}

Figure~\ref{fig:commercial_prompt} shows the prompt template used for all commercial models.

\subsection{Qualitative Comparison}

Figures~\ref{fig:commercial1} and Figures~\ref{fig:commercial2} compare GPT-5.4 Pro, Nano Banana Pro, and AsemConsist on three representative prompt sets. All methods receive identical identity prompts and image-specific prompts under the protocol described above.

\subsection{Practical Usage}

For reference, commercial image generation incurs a recurring API cost for every generated sequence.

\begin{itemize}
    \item \textbf{GPT-Image-2 medium:} approximately \$0.108 per four-image sequence.
    \item \textbf{Nano Banana Pro:} approximately \$0.1417 per four-image sequence.
\end{itemize}

Evaluating our 90-sequence benchmark therefore costs approximately \$9.72 with GPT-Image-2 medium and \$12.75 with Nano Banana Pro. Although generating a single image sequence is affordable, practical applications such as storyboards and long-form visual narratives often require hundreds or thousands of generation requests. As a result, both API cost and generation latency accumulate over iterative content creation. In contrast, AsemConsist operates locally on open-weight diffusion models without recurring per-generation API charges.

\section{Hyperparameter Sensitivity}
\label{app:sensitivity}

We evaluate the sensitivity of the inference-time hyperparameters in STM and PAD. Starting from the default configuration ($\alpha_{\exp}=0.025$, $\alpha_{\sup}=-0.01$, and $\gamma=0.85$), we vary one hyperparameter at a time while keeping the others fixed. Tables~\ref{tab:sensitivity-svr} and~\ref{tab:sensitivity-pad} report the results.

Across both FLUX and SD3.5, STM is insensitive to the rescaling strengths ($\alpha_{\exp}, \,\, \alpha_{\sup}$), suggesting that the adaptive component-selection strategy is the dominant factor, while the exact rescaling magnitude is less critical. Similarly, PAD remains stable across different blend weights, indicating that semantic projection provides a robust mechanism for injecting expression semantics into the padding embeddings.

\section{Choice of Spectral Decomposition}
We use singular value decomposition (SVD) to manipulate text embeddings through their dominant spectral subspaces~\cite{1p1s}, applying it directly to the uncentered embedding matrix so that the token-wise mean direction is preserved for resolving the SVD sign ambiguity~\cite{bro2008resolving}. As the embedding dimension is small, exact SVD is cheap yet stable. PCA is a closely related alternative and would likely yield similar results.
\clearpage

\begin{table*}[t]
\centering

\begingroup
\renewcommand{\arraystretch}{0.95}
\setlength{\tabcolsep}{4.5pt}

\small
\begin{tabular*}{\textwidth}{lcc@{\extracolsep{\fill}}ccc@{\extracolsep{\fill}}ccc}
\toprule
& & &
\multicolumn{3}{c}{FLUX} &
\multicolumn{3}{c}{SD3.5} \\
\cmidrule(lr){4-6}
\cmidrule(lr){7-9}

Parameter & Scale & Value
& SeeSaw$\uparrow$
& Text alignment$\uparrow$
& DreamSim$\downarrow$
& SeeSaw$\uparrow$
& Text alignment$\uparrow$
& DreamSim$\downarrow$ \\

\midrule

\multicolumn{3}{l}{\textit{Vanilla}}
&0.584&0.723&0.347
&0.636&0.820&0.367\\

\midrule

\multicolumn{3}{l}{\textbf{Default}}
&\textbf{0.664}
&\textbf{0.786}
&\textbf{0.281}
&\textbf{0.681}
&\textbf{0.809}
&\textbf{0.272}\\

\midrule

$\alpha_{\exp}$
&0.5$\times$
&0.0125
&0.671&0.791&0.279
&0.678&0.812&0.276\\

&0.75$\times$
&0.01875
&0.668&0.780&0.278
&0.673&0.806&0.276\\

&1.5$\times$
&0.0375
&0.652&0.764&0.283
&0.673&0.793&0.275\\

&2$\times$
&0.0500
&0.650&0.752&0.282
&0.670&0.820&0.284\\

\midrule

$\alpha_{\sup}$
&0.5$\times$
&-0.005
&0.667&0.785&0.279
&0.679&0.803&0.273\\

&0.75$\times$
&-0.0075
&0.663&0.779&0.281
&0.682&0.809&0.272\\

&1.5$\times$
&-0.015
&0.662&0.780&0.281
&0.678&0.804&0.273\\

&2$\times$
&-0.020
&0.661&0.781&0.283
&0.682&0.809&0.272\\

\bottomrule
\end{tabular*}
\endgroup

\caption{Sensitivity of STM to the SVR rescaling strengths. The default hyperparameters used throughout the paper are shown separately. All remaining rows vary a single parameter by $0.5\times$--$2\times$ while keeping the others fixed.}
\label{tab:sensitivity-svr}
\end{table*}
\begin{table*}[!t]
\centering

\begingroup
\renewcommand{\arraystretch}{0.95}
\setlength{\tabcolsep}{4.5pt}

\small
\begin{tabular*}{\textwidth}{lc@{\extracolsep{\fill}}ccc@{\extracolsep{\fill}}ccc}
\toprule
& &
\multicolumn{3}{c}{FLUX} &
\multicolumn{3}{c}{SD3.5} \\
\cmidrule(lr){3-5}
\cmidrule(lr){6-8}

Setting & $\gamma$
& SeeSaw$\uparrow$
& Text alignment$\uparrow$
& DreamSim$\downarrow$
& SeeSaw$\uparrow$
& Text alignment$\uparrow$
& DreamSim$\downarrow$ \\

\midrule

Vanilla & --
& 0.584 & 0.723 & 0.347
& 0.636 & 0.820 & 0.367 \\

\textbf{Default} & \textbf{0.85}
& \textbf{0.664} & \textbf{0.786} & \textbf{0.281}
& \textbf{0.681} & \textbf{0.809} & \textbf{0.272} \\

\midrule

PAD & 0.65
& 0.668 & 0.781 & 0.278
& 0.680 & 0.807 & 0.273 \\

& 0.75
& 0.664 & 0.779 & 0.280
& 0.678 & 0.811 & 0.274 \\

& 0.95
& 0.658 & 0.752 & 0.276
& 0.678 & 0.802 & 0.272 \\

\bottomrule
\end{tabular*}
\endgroup

\caption{Sensitivity of PAD to the semantic-projection blend weight $\gamma$. The default value used throughout the paper is shown separately, and all other rows vary only $\gamma$ while keeping the remaining settings fixed.}
\label{tab:sensitivity-pad}
\end{table*}

\end{document}